\documentclass[runningheads]{llncs}

\usepackage{eccv}

\usepackage{eccvabbrv}

\usepackage{graphicx}
\usepackage{booktabs}
\usepackage{subcaption}
\usepackage{amssymb}
\usepackage{tabularx} 
\usepackage{array}
\usepackage{multirow, multicol}
\usepackage{caption}
\usepackage{subcaption}
\usepackage{color}
\usepackage{comment}

\usepackage[accsupp]{axessibility}  

\usepackage[pagebackref,breaklinks,colorlinks]{hyperref}

\usepackage{orcidlink}
\usepackage{enumitem}

\usepackage{xcolor} 

\definecolor{darkergreen}{RGB}{0,170,0} 

\newcommand{\red}[1]{{\color{red}#1}}
\newcommand{\blue}[1]{{\color{blue}#1}}
\newcommand{\green}[1]{{\color{darkergreen}#1}}

\newcommand{\ourframework}{Scene Flow via Tracking}
\newcommand{\oureval}{Bucket Normalized EPE}

\newcommand{\ourmethod}{TrackFlow}
\newcommand{\ourmethodbevfusion}{\ourmethod{}BEVF}

\newcommand{\zerovec}{\overrightarrow{0}}

\newcommand{\figref}[1]{Fig.~\ref{fig:#1}}

\newcommand{\figlabel}[1]{\label{fig:#1}} 
\newcommand{\tableref}[1]{Table~\ref{table:#1}}

\newcommand{\tablelabel}[1]{\label{table:#1}}

\newcommand{\sectionref}[1]{Section~\ref{section:#1}}
\newcommand{\sectionlabel}[1]{\label{section:#1}}
\newcommand{\appendixref}[1]{Appendix~\ref{appendix:#1}}
\newcommand{\appendixlabel}[1]{\label{appendix:#1}}
\newcommand{\equationref}[1]{Equation~\ref{equation:#1}}
\newcommand{\equationlabel}[1]{\label{equation:#1}}

\newcommand{\pointcloudt}{P_t}
\newcommand{\pointcloudtpone}{P_{t+1}}

\newcommand{\flowttpone}{\hat{F}_{t,t+1}}

\newcommand{\flowgtttpone}{F^*_{t,t+1}}
\newcommand{\norm}[1]{\left\lVert #1 \right\rVert}

\newcommand{\tinier}{\fontsize{2pt}{2pt}\selectfont}

\begin{document}

\title{\emph{I Can't Believe It's Not Scene Flow!}}



\author{Ishan Khatri$^{1,3*}$, Kyle Vedder$^{2*}$, Neehar Peri$^3$, Deva Ramanan$^3$, James Hays$^4$}

\authorrunning{I.~Khatri et al.}

\institute{$^1$Stack AV, $^2$University of Pennsylvania, $^3$CMU, $^4$Georgia Tech}

\maketitle

\vspace{-1em}
\begin{figure}[h!]
\centering
\begin{subfigure}{.32\textwidth}
  \centering
  \includegraphics[width=\linewidth]{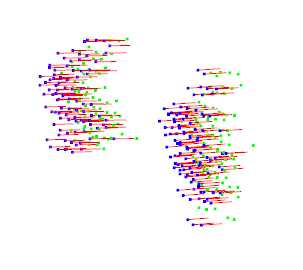}
  \caption{Ground Truth}
  \label{fig:sub5}
\end{subfigure}%
\begin{subfigure}{.32\textwidth}
  \centering
  \includegraphics[width=\linewidth]{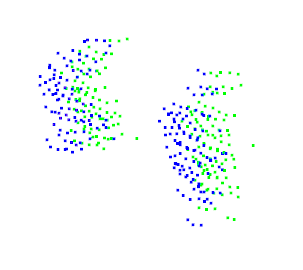}
  \caption{FastFlow3D~\cite{scalablesceneflow}}
  \label{fig:sub2}
\end{subfigure}
\begin{subfigure}{.32\textwidth}
  \centering
  \includegraphics[width=\linewidth]{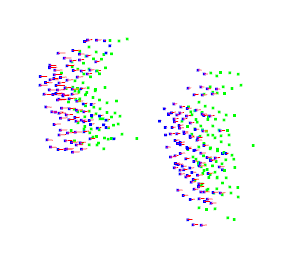}
  \caption{DeFlow~\cite{zhang2024deflow}}
  \label{fig:sub3}
\end{subfigure}

\begin{subfigure}{.32\textwidth}
  \centering
  \includegraphics[width=\linewidth]{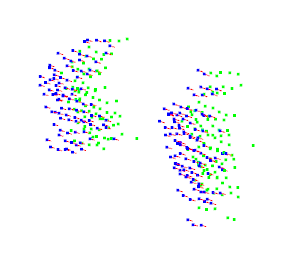}
  \caption{NSFP~\cite{nsfp}}
  \label{fig:sub4}
\end{subfigure}%
\begin{subfigure}{.32\textwidth}
  \centering
  \includegraphics[width=\linewidth]{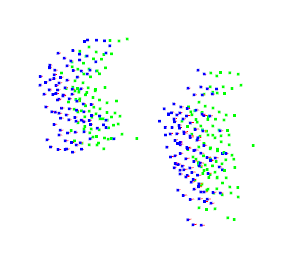}
  \caption{ZeroFlow XL 5x~\cite{vedder2024zeroflow}}
  \label{fig:sub1}
\end{subfigure}%
\begin{subfigure}{.32\textwidth}
  \centering
  \includegraphics[width=\linewidth]{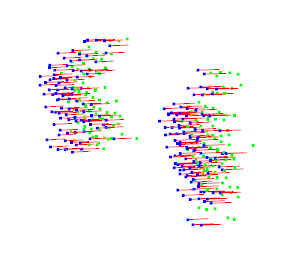}
  \caption{\textbf{\ourmethod{} (Ours)}}
  \label{fig:sub6}
\end{subfigure}
\caption{We visualize an example of two pedestrians (walking from \blue{left} to \green{right}), cherry-picked to have unusually high density lidar returns, making it particularly easy to estimate flow. We expect that state-of-the-art scene flow methods should work well in this case, but find that all prior art fails catastrophically. Notably, \ourmethod{} is the only method to estimate flow for these pedestrians.}
\figlabel{teaserfigure}
\vspace{-1cm}
\end{figure} 

\begin{abstract}
State-of-the-art scene flow methods broadly fail to describe the motion of small objects, and existing evaluation protocols hide this failure by averaging over many points. To address this limitation, we propose \emph{\oureval{}}, a new class-aware and speed-normalized evaluation protocol that better contextualizes error comparisons between object types that move at vastly different speeds. In addition, we propose \emph{\ourmethod{}}, a frustratingly simple supervised scene flow baseline that combines a high-quality 3D object detector (trained using standard class re-balancing techniques) with a simple Kalman filter-based tracker. Notably, \emph{\ourmethod{}} achieves state-of-the-art performance on existing metrics and shows large improvements over prior work on our proposed metric. Our results highlight that scene flow evaluation must be class and speed aware, and supervised scene flow methods must address point-level class imbalances. Our evaluation toolkit and code is available on \href{https://github.com/kylevedder/BucketedSceneFlowEval}{GitHub}.


\end{abstract}

\section{Introduction}

Scene flow estimation is the task of describing a 3D motion field between temporally successive point clouds~\cite{vedula1999, Zhai2020FlowMOT3M, baur2021slim, flownet3d, flowssl, scalablesceneflow, vedder2024zeroflow}. In theory, high quality flow estimators can provide a valuable signal about scene-level dynamics~\cite{scalablesceneflow, vedder2024zeroflow} 
for both online~\cite{Zhai2020FlowMOT3M} and offline~\cite{objectdetectionmotion} processing. Do state-of-the-art scene flow methods actually work well in practice? 

{\bf Status Quo.} Standard scene flow metrics suggest that existing methods can estimate motion to centimeter-level accuracy. 
For example, ZeroFlow XL 5x~\cite{vedder2024zeroflow} achieves 
an average Threeway EPE~\cite{chodosh2023} of only 4.9 centimeters (1.9 inches) and a Dynamic EPE (averaged over points moving faster than 0.5 m/s)  of 11.7 centimeters (4.6 inches). Notably, these errors are relatively small compared to the scale of cars and pedestrians, implying that current scene flow methods produce high quality flow. On the scale of cars and people, these \emph{feel} like tiny errors and seem to imply that current scene flow methods are high quality.

{\bf \oureval{}.}  We visualize flow predictions from several state-of-the-art supervised (FastFlow3D~\cite{scalablesceneflow}, DeFlow~\cite{zhang2024deflow}) and unsupervised (NSFP~\cite{nsfp}, ZeroFlow~\cite{vedder2024zeroflow}) approaches and find that all methods underestimate flow for small objects (\figref{teaserfigure}) with fewer lidar points (e.g. pedestrians and bicyclists). 
Surprisingly, existing scene flow metrics do not highlight such failure cases on these safety-critical categories because small objects only make up a tiny fraction of the dynamic points in a scene (\figref{fig:pointdistribution}).
To address this limitation, we propose \emph{\oureval{}}, a new evaluation protocol that allows us to directly measure performance disparities across classes of different sizes and speed profiles. Specifically, \emph{\oureval{}} evaluates the \emph{percentage} of described motion, allowing us to normalize comparisons between objects moving at different speeds. 
Our proposed evaluation metric takes inspiration from \emph{mean Average Precision} (mAP), a metric commonly used to evaluate object detectors. Notably, unlike existing scene flow metrics, mAP equally weights the performance of large common objects like cars and small rare objects like strollers. Therefore, state-of-the-art 3D object detectors use data augmentation and class re-balancing techniques~\cite{cbgs} to perform well on both common and rare classes. 

{\bf \ourmethod{}.} Based on this observation, we propose \ourmethod{}, a frustratingly simple baseline that generates scene flow estimates using rigid transformations to describe point-level motion within a 3D object track.  Specifically, we run a state-of-the-art 3D object detector~\cite{wang2023le3de2e} followed by a simple 3D Kalman filter-based tracker~\cite{Weng2020_AB3DMOT} to generate object trajectories. 
Despite its simplicity,  \emph{\ourmethod{}} achieves state-of-the-art performance on Threeway EPE and significantly outperforms prior art on our \oureval{} metric, capturing an additional 10\% of total motion in general and an additional 20\% of total motion on pedestrians (a $1.5\times$ improvement). Importantly, our simple baseline's state-of-the-art performance is an indictment of existing supervised scene flow methods. We argue that utilizing (well established) class re-balancing techniques can improve performance on rare safety-critical categories in real-world datasets, and evaluating scene flow methods using class and speed-aware metrics more closely reflects real-world performance. 

{\bf Contributions}. We present three primary contributions.
\begin{enumerate}
  \item We highlight the qualitative failure of state-of-the-art scene flow methods on safety-critical categories like pedestrians and bicycles.
  \item We introduce \emph{\oureval{}}, a new evaluation protocol that allows us to quantify this qualitative failure on small objects.
  \item We propose \emph{\ourmethod{}}, a frustratingly simple baseline that achieves state-of-the-art performance on standard metrics and significantly outperforms prior art on our class-aware \emph{\oureval{}} metric.
\end{enumerate}

\section{Related Work}

\subsection{Scene Flow Datasets and Ground Truth}\sectionlabel{sceneflowdatasets}
Unlike next token prediction in language~\cite{gpt} or next frame prediction in vision~\cite{weng2021inverting}, scene flow is not na\"ively self-supervised: future observations do not provide ground truth scene flow. Therefore, ground truth motion descriptions must be provided by an oracle, typically from human annotators for real data~\cite{kittisceneflow1, kittisceneflow2, waymoopen, argoverse2, caesar2020nuscenes} or a data generator for synthetic datasets~\cite{flyingthings,zheng2023point}. For real world datasets (typically from the autonomous vehicle domain) human annotations are provided in the form of 3D bounding boxes and tracks for every object in the scene~\cite{chodosh2023}. Consequently, the generated ground truth flow is assumed to be rigid, even in the case of non-rigid motion like pedestrian gaits.

\subsection{Scene Flow Estimation}
Given point clouds $\pointcloudt{}$ and $\pointcloudtpone{}$, scene flow estimators predict $\flowttpone{}$,  a 3D vector per point in $\pointcloudt$ that describes its motion from $t$ to $t+1$ \cite{dewan2016rigid}. Performance is typically measured using Average Endpoint Error (EPE) which is the $L_2$ norm between the predicted ($\flowttpone$) and ground truth flow ($\flowgtttpone$), as in \equationref{averageepedef}.

\begin{equation}
  \small
  \equationlabel{averageepedef}
  \textup{Average EPE}\left({\pointcloudt} \right) = \frac{1}{\norm{\pointcloudt}} \sum_{p \in \pointcloudt} \norm{\flowttpone{}(p) - \flowgtttpone{}(p)}_2.
\end{equation}

Current state-of-the-art methods for scene flow estimation broadly fall into one of two categories: supervised and unsupervised.

\subsubsection{Supervised Scene Flow} methods train feedforward networks to perform flow vector regression based on ground truth annotations~\cite{flownet3d,behl2019pointflownet,tishchenko2020self,kittenplon2021flowstep3d,wu2020pointpwc,puy2020flot,li2021hcrf,scalablesceneflow,gu2019hplflownet,battrawy2022rms, 9856954, zhang2024deflow,li2018pointcnn}. Many of these networks utilize custom point operations such as point-based convolutions~\cite{flownet3d,gu2019hplflownet, kittenplon2021flowstep3d,li2018pointcnn}, making them intractable to train on large point clouds. In contrast, FastFlow3D~\cite{scalablesceneflow} uses a feedforward architecture based on PointPillars~\cite{pointpillars}, an efficient lidar detector architecture, to train and predict flow on real-world large-scale point clouds. FastFlow3D's speed and quality make it a popular base architecuture for both unsupervised and supervised methods like ZeroFlow~\cite{vedder2024zeroflow} and DeFlow~\cite{zhang2024deflow}, respectively.

\subsubsection{Unsupervised Scene Flow} methods tend to use online optimization against surrogate objectives such as Chamfer distance\cite{nsfp}, cycle-consistency \cite{Mittal_2020_CVPR}, distance transforms \cite{Li_2023_ICCV}, or other hand-designed heuristics \cite{chodosh2023, pontes2020scene, gojcic2021weakly}. For example, Neural Scene Flow Prior (NSFP)~\cite{nsfp} provides high quality scene flow estimates by optimizing a small ReLU MLP at test time to minimize Chamfer distance and maintain cycle-consistency.
Other unsupervised methods like ZeroFlow~\cite{vedder2024zeroflow} indirectly leverage online optimization. Vedder et. al ~\cite{vedder2024zeroflow} introduces \emph{Scene Flow via Distillation}, a framework that uses a slow optimization-based method to pseudolabel unlabeled point cloud pairs and trains a fast feedforward network with these pseudolabels.

\subsection{Scene Flow Evaluation Metrics}
In real-world scenes, most points belong to the static background. Consequently, simply computing Average EPE (\equationref{averageepedef}) over all points is dominated by background points. 
In order to separately measure non-ego dynamics, Chodosh et al.~\cite{chodosh2023} introduces \emph{Threeway EPE}, which computes a mean over the Average EPE for three disjoint classes of points: \emph{Foreground Dynamic} (points inside bounding box labels moving greater than 0.5m/s), \emph{Foreground Static} (points inside bounding box labels moving less than 0.5m/s), and \emph{Background Static}. We extend Threeway EPE to consider different class and speed profiles.

\subsection{3D Object Detection and 
Tracking}\sectionlabel{detectorrelatedwork}

Object detectors have advanced techniques for training with imbalanced datasets. Notably, modern object detectors use carefully designed losses to mitigate foreground-vs-background imbalances in proposal generation, and data augmentation strategies to train with long-tailed taxonomies. Existing methods address imbalanced foreground-vs-background region proposals using Focal Loss~\cite{focalloss} to upweight the importance of foreground regions. More recently, 3D object detectors use class-balanced sampling~\cite{cbgs} and copy-paste augmentation to upsample and rebalance the distribution of examples per class. In addition, state-of-the-art 3D object detectors take advantage of multi-modal data to improve detection \cite{vora2020pointpainting, peri2022towards, ma2023long} of small and rare categories. Since many state-of-the-art tracking algorithms \cite{Weng2020_AB3DMOT} follow the tracking-by-detection paradigm, improving detection quality also significantly improves tracking performance.

\section{\emph{\oureval{}}: Small Objects (Should) Matter in Scene Flow}\sectionlabel{eval}

As shown in \figref{teaserfigure} (and further in \figref{morequalitativeone}), existing scene flow methods consistently struggle to describe the motion of safety-critical objects like pedestrians. However, these failures are not captured by Threeway EPE \emph{because} these objects are small and have few points. Specifically, Threeway EPE's \emph{Foreground Dynamic} category is dominated by large, common objects with many points like cars and other vehicles. As shown in \figref{fig:pointdistribution}, 15\% of all points are from cars or other vehicles (dominating \emph{Foreground Dynamic}'s Average EPE), while fewer than 1\% of points are from pedestrians and other vulnerable road users (VRUs). 

\begin{figure}[t]
\centering
\includegraphics{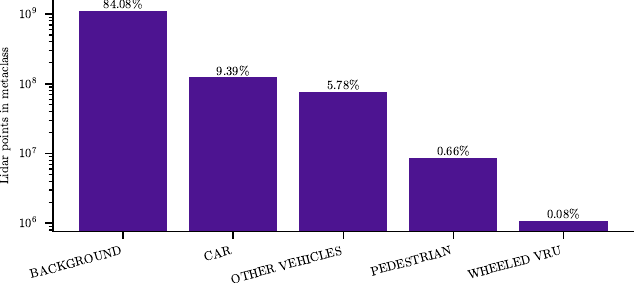}
\caption{Number of points from each semantic meta-class for Argoverse 2's \emph{val} split. Although \texttt{PEDESTRIAN} instances are common, they contribute less than 1\% of the total number of points owing to their small instance size relative to \texttt{CAR} and \texttt{OTHER VEHICLES}. Number of points (Y axis) shown on a log scale.}
\figlabel{fig:pointdistribution}
\end{figure}

Additionally, Threeway EPE fails to account for large differences in speed  across objects. For example, a 0.5m/s estimation error on a car moving 20 m/s is negligible (<2.5\%), while a 0.5m/s estimation error on a pedestrian moving 0.5m/s fails to describe 100\% of the pedestrian's motion. However, Threeway EPE treats both estimation errors equally.

We address these two limitations with our \emph{\oureval{}} metric. First, our proposed metric breaks down the object distribution using a taxonomy that human labelers have deemed important (similar to \emph{mean Average Precision}~\cite{cocodataset}, see \appendixref{semanticsfree} for discussion on semantics-free evaluation). Second, our proposed metric allows us to contextualize the percentage of object motion being described by normalizing for the speed of the object, allowing us to directly compare performance across object categories.


{
\setlength{\tabcolsep}{1.4em}

\begin{table}[t]
\centering
\scalebox{0.99}{
\begin{tabular}{lcc}
\toprule
Class & Static (Avg EPE) & Dynamic (Norm EPE) \\
\midrule
\texttt{BACKGROUND} & 0.002402 & - \\
\texttt{CAR} & 0.018442 & 0.182092 \\
\texttt{OTHER VEHICLES} & 0.081475 & 0.312882 \\
\texttt{PEDESTRIAN} & 0.052842 & 0.396849 \\
\texttt{WHEELED VRU} & 0.062573 & 0.257647 \\
\bottomrule
\end{tabular}
}
\vspace{5mm}
\caption{\ourmethod{}'s \oureval{} on the Argoverse 2 \emph{test} split. Similar to Threeway EPE, we breakdown our evaluation into static and dynamic buckets. However, we also further breakdown performance by meta-categories and normalize by speed to compare performance disparities on safety-critical categories. TrackFlow is able to capture most dynamic car motion (lower is better), but performs considerably worse on other vehicles and pedestrian.}
\tablelabel{tab:tracktorflowresults}
\end{table}
}

We implement our class-aware and speed-normalized metric by accumulating every point into a class-speed matrix (e.g.\ \appendixref{ourevalstructure}, \tableref{speedclassmatrix}) based on its ground truth speed and class, recording an Average EPE as well as a per-bucket average speed. To summarize these results, we report two numbers per class:

\begin{itemize}
    \item \emph{Static EPE}, taken directly from the Average EPE of the first speed bucket for that class (i.e.\ the first column of \appendixref{ourevalstructure}, \tableref{speedclassmatrix})
    \item \emph{Dynamic Normalized EPE}, computed from a mean over the Normalized EPE ($\frac{\textup{Average EPE}}{\textup{average speed}}$) of each non-empty speed bucket (i.e.\ an average across the Normalized EPEs of the second column onwards in \appendixref{ourevalstructure}, \tableref{speedclassmatrix})
\end{itemize}

\emph{Dynamic Normalized EPE} measures the fraction of motion {\em not} described by the estimated flow vectors across the entire speed spectrum. A method that only predicts ego motion (e.g.\ $\zerovec$ after ego-motion compensation) will achieve 1.0 Dynamic Normalized EPE, and a method that perfectly describes all motion will have 0.0 Dynamic Normalized EPE. Methods may achieve errors greater than 1.0 by predicting errors with a magnitude greater than the average speed. For example, a method that describes the negative vector of true motion will get exactly 2.0 Dynamic Normalized EPE (every bucket's Average EPE will be exactly 2$\times$ the magnitude of the average speed). The range of Dynamic Normalized EPE is between 0 (perfect) and $\infty$, and is undefined for buckets without any points. After normalization, Dynamic Normalized EPE can be directly compared across classes.

We provide an example per-class performance breakdown in \tableref{tab:tracktorflowresults} for \ourmethod{} (\sectionref{method}). Results can be further summarized into a single tuple of \emph{mean Static EPE} and \emph{mean Dynamic Normalized EPE} by taking a mean across classes (similar to \emph{mean Average Precision}~\cite{cocodataset}). \ourmethod{} has a mean Static EPE of 0.076277 and a mean Dyanmic Normalized EPE of 0.287368. We rank methods according to their mean Dynamic Normalized EPE.

\section{\emph{\ourmethod{}}: \ourframework{}}\sectionlabel{method}

 To highlight the failure of current supervised scene flow methods on smaller objects, we propose \emph{\ourframework{}}, a simple framework that uses bounding box track motion from off-the-shelf 3D detectors and trackers to generate scene flow estimates (\figref{fig:detectplustrack}).
 We instantiate \emph{\ourframework{}} with LE3DE2E~\cite{wang2023le3de2e}\footnote{LE3DE2E~\cite{wang2023le3de2e} is the winning method from the \emph{Argoverse 2 2023 3D Detection, Tracking and Forecasting challenge}~\cite{peri2022towards, peri2023empirical, Peri_2022_CVPR}.}, a state-of-the-art 3D detector, and AB3DMOT~\cite{Weng2020_AB3DMOT}, a Kalman filter-based 3D tracker. As shown in \sectionref{experiments}, \emph{\ourmethod{}} achieves state-of-the-art performance on Threeway EPE and beats all prior art by a large margin on \oureval{}.

\begin{figure}[t]
    \centering
    \includegraphics[width=\textwidth]{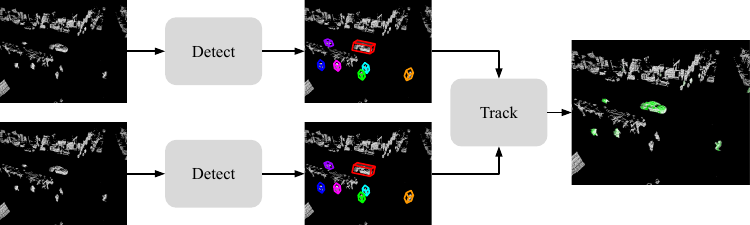}
    \caption{Overview of the \emph{\ourframework{}} framework. Our proposed framework generates scene flow estimates using rigid transformations to describe point-level motion within a 3D object track.}
    \figlabel{fig:detectplustrack}
\end{figure}

\ourframework{} works well in practice because it mimics the ground truth flow annotation process. Specifically, ground truth flow is generated using rigid transforms to describe point-level motion within ground truth 3D object tracks
(\sectionref{sceneflowdatasets}). Therefore, a perfect 3D detector and tracker will achieve perfect flow. However, the power of \ourmethod{} isn't just derived from its use of bounding boxes; it also greatly benefits from recent advances in class-imbalanced learning~\cite{cbgs}. As discussed in \sectionref{detectorrelatedwork}, modern detectors are trained with a variety of data augmentation techniques to achieve high precision and recall on all semantic class. \ourmethod{} leverages the strength of modern 3D detectors to significantly outperform prior art on pedestrians and other small objects.

Interestingly, we find that the \ourframework{} framework performs best when using detectors tuned to a low confidence threshold. Typically, detectors are optimized to only predict high confidence boxes (0.7 - 0.9) to minimize the number of false positives. However, our method works best when setting the confidence threshold lower (0.2 for \ourmethod{}) to increase recall. Specifically, we find that detectors with higher recall and more accurate heading estimation are better suited for \ourframework{}. We explore detector choice and ablate the impact of confidence thresholds further in \sectionref{detectorquality}.

\section{Experiments}\sectionlabel{experiments}

In this section, 
we compare \ourmethod{} against state-of-the-art supervised and unsupervised scene flow methods like FastFlow3D~\cite{scalablesceneflow}, DeFlow~\cite{zhang2024deflow}, NSFP~\cite{nsfp}, and ZeroFlow~\cite{vedder2024zeroflow} on the Argoverse 2 benchmark~\cite{argoverse2}\footnote{All evaluations are performed with a maximum radius of 35m from the ego vehicle to maintain consistency with Chodosh et al.~\cite{chodosh2023}.}.

\begin{figure}[t]
\centering
\begin{subfigure}[b]{0.49\textwidth}
    \includegraphics{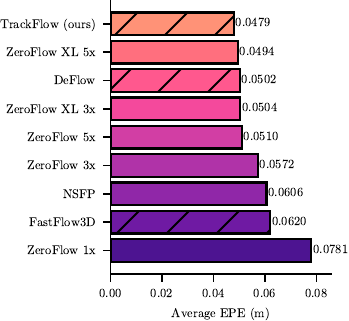}  
    \caption{Threeway EPE}
    \figlabel{fig:threewayepesub}
\end{subfigure}%
\begin{subfigure}[b]{0.49\textwidth}
    \includegraphics{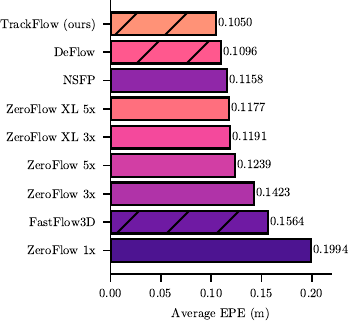}    
    \caption{Threeway EPE's \emph{Foreground Dynamic}}
    \figlabel{fig:threewayepedynamicsub}
\end{subfigure}%
\caption{\emph{Threeway EPE} and \emph{Threeway EPE's Foreground Dynamic} performance of recent supervised and unsupervised scene flow  methods on Argoverse 2's \emph{test} split. Supervised methods shown with hatching. Lower is better. Method color is consistent between plots. We find that all recent methods achieve 5cm error on Threeway EPE, suggesting that these approaches work well in-the-wild. However, this number hides the failure of these methods to describe small object motion.}
\figlabel{fig:threewayepe}
\end{figure}

\begin{figure}[t]
\centering
\begin{subfigure}[b]{0.49\textwidth}
    \centering
    \includegraphics{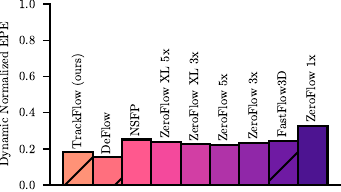}
    \caption{\texttt{CAR}}
    \figlabel{fig:car}
\end{subfigure}%
\begin{subfigure}[b]{0.49\textwidth}
    \centering
    \includegraphics{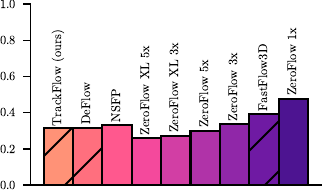}
    \caption{\texttt{OTHER VEHICLES}}
    \figlabel{fig:other-vehicles}
\end{subfigure}
\begin{subfigure}[b]{0.49\textwidth}
    \centering
    \includegraphics{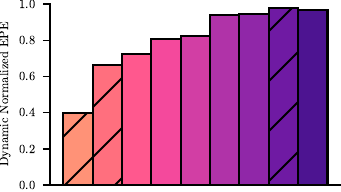}
    \caption{\texttt{PEDESTRIAN}}
    \figlabel{fig:pedestrian}
\end{subfigure}%
\begin{subfigure}[b]{0.49\textwidth}
    \centering
    \includegraphics{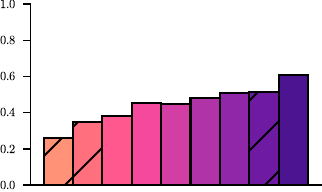}
    \caption{\texttt{WHEELED VRU}}
    \figlabel{fig:wheeled-vru}
\end{subfigure}
\caption{Per meta-class Dynamic Normalized EPE of recent supervised and unsupervised scene flow estimation methods on Argoverse 2's \emph{test} split. Supervised methods shown with hatching. Lower is better. Method color and position is consistent between plots. \ourmethod{} significantly improves over prior work on both pedestrian and wheeled VRUs. Notably, \oureval{} quantitatively demonstrates significant method performance differences not highlighted in Threeway EPE.}
\figlabel{metacatagorydynamic}
\end{figure}

\begin{figure}[t]
\centering
\includegraphics{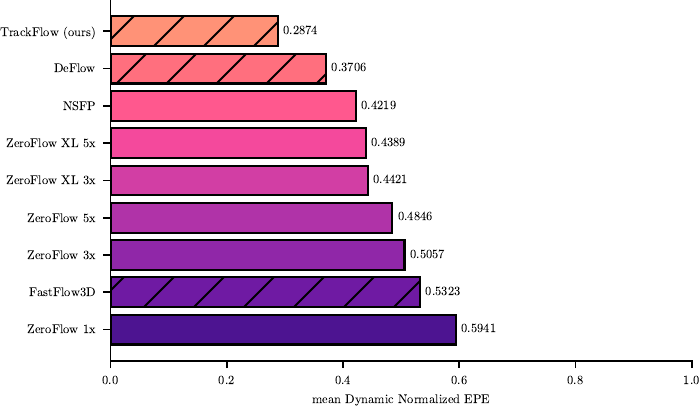}
\caption{Average Dynamic Normalized EPE of recent supervised and unsupervised scene flow estimation methods on Argoverse 2's \emph{test} split. Supervised methods shown with hatching. Lower is better. Our simple baseline achieves state-of-the-art performance, suggesting that supervised scene flow methods should embrace point-level class re-balancing.}
\figlabel{fig:meandynamicepe}
\end{figure}

{
\setlength{\tabcolsep}{1.4em}

\begin{table}[t]
\centering
\scalebox{0.99}{
\begin{tabular}{lcc}
\toprule
Class & Static (Avg EPE) & Dynamic (Norm EPE) \\
\midrule
\texttt{BACKGROUND} & \green{-0.000228} & - \\
\texttt{CAR} & \red{+0.039049} & \red{+0.117944} \\
\texttt{OTHER VEHICLES} & \red{+0.009013} & \red{+0.224830} \\
\texttt{PEDESTRIAN} & \red{+0.007187} & \red{+0.224250} \\
\texttt{WHEELED VRU} & \green{-0.025889} & \red{+0.151373} \\
\bottomrule
\end{tabular}
}
\vspace{5mm}
\caption{Relative \oureval{} performance of \ourmethodbevfusion{} compared to \ourmethod{}, on the Argoverse 2's \emph{test} split. Increases in error (worse) are shown with a + in \red{red}, and decreases in error (better) are shown with a - in \green{green}. \ourmethod{}'s absolute results are shown in \tableref{tab:tracktorflowresults}.BEVFusion only has 2\% lower mAP than LE3DE2E on the AV2 detection leaderboard, but performs significantly worse than \ourmethod{} on Dynamic Normalized EPE.}
\tablelabel{tab:bevfusionresults}
\end{table}
}

{
\setlength{\tabcolsep}{3.35em}

\begin{table}[htb]
\centering
\scalebox{0.99}{
\begin{tabular}{cc}
\hline
Confidence & Mean Dynamic Norm EPE \\ \hline
0.1        & 0.4816             \\
0.2        & 0.4643             \\
0.3        & 0.6008             \\
0.4        & 0.8176             \\ \hline
\end{tabular}
}
\vspace{5mm}
\caption{Mean Dynamic Bucketed EPE values for TrackFlowBEVF using various confidence thresholds for the detector. Lower confidences with higher recall significantly improve Dynamic Norm EPE performance.}
\tablelabel{tab:bevfusionconf}
\end{table}
}

\begin{figure}[h!]
    \centering
    \includegraphics[width=\textwidth]{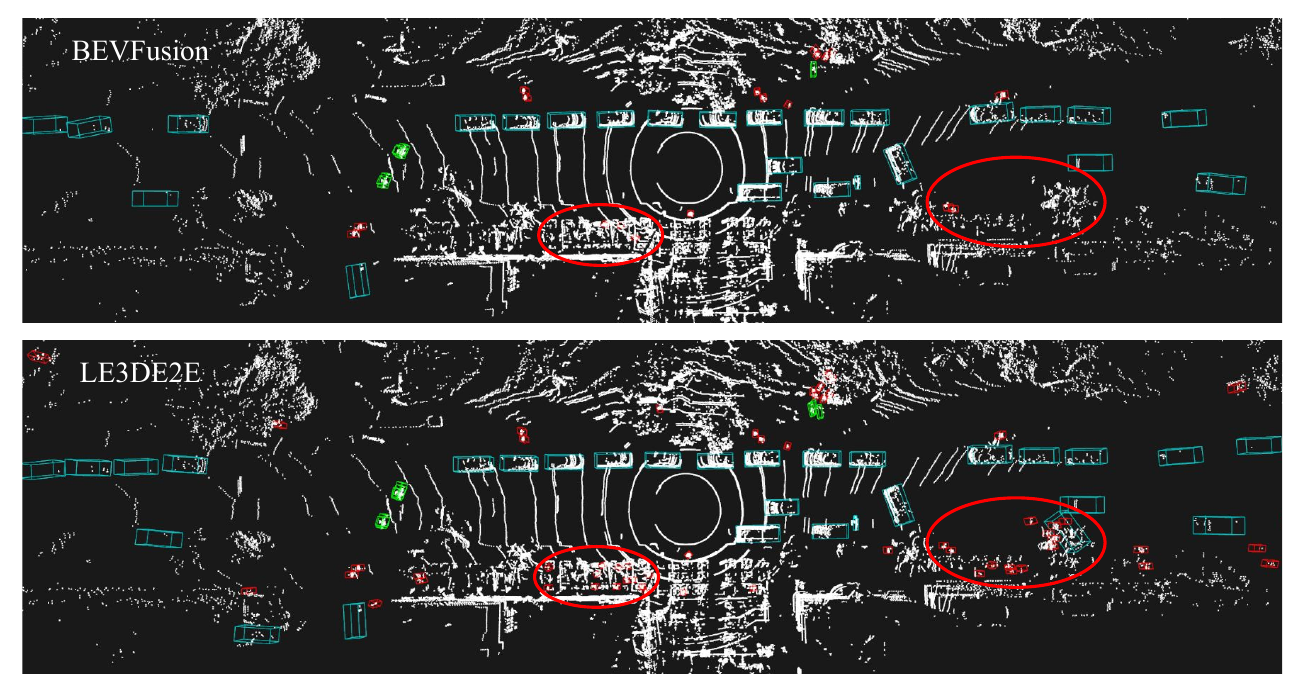}
    \caption{A qualitative comparison of the recall of BEVFusion and LE3DE2E. LE3DE2E has much higher recall, allowing it to pick out pedestrians BEVFusion missed (circled in red), and better quality box heading estimates. Both detectors are using a confidence threshold of 0.2.}
    \figlabel{fig:bevfusionrecall}
    \vspace{-2em}
\end{figure}

\subsection{\emph{\ourmethod{}} Achieves SOTA Performance on Threeway EPE}

\ourmethod{} is state-of-the-art on Threeway EPE (\figref{fig:threewayepesub}) on the Argoverse 2 benchmark~\cite{argoverse2}, achieving an overall reduction of 0.0015m (0.15cm, or 1.5mm) over the next best method, ZeroFlow XL 5x. Notably, this improvement can be attributed to significantly better Threeway EPE's \emph{Dynamic Foreground} (\figref{fig:threewayepedynamicsub}). Is this performance difference meaningful? 

Based on our reduction of 1.5mm on Threeway EPE (about 4$\times$ the thickness of a human fingernail), it would seem that \ourmethod{} is only an incremental improvement over prior art. However, \ourmethod{} qualitatively outperforms prior work on important small objects such as pedestrians (\figref{teaserfigure}, \figref{morequalitativeone}). As shown in the next section, our proposed evaluation protocol \emph{\oureval{}} makes it quantitatively clear that \ourmethod{} performs significantly better on safety-critical categories like pedestrians and VRUs.


\subsection{\emph{\oureval{}} Highlights Failures on Small Objects}

Evaluating existing state-of-the-art methods on our class-aware, speed-normalized evaluation, \oureval{}, makes it clear that \ourmethod{} meaningfully outperforms prior art (\figref{fig:meandynamicepe}) --- \ourmethod{} correctly describes almost 10\% additional total motion across meta-classes compared to  DeFlow~\cite{zhang2024deflow}. This difference in dynamic performance becomes even more clear when broken down by meta-class: \figref{metacatagorydynamic} shows that \ourmethod{} is the only method able to describe more than 50\% of pedestrian motion, beating DeFlow~\cite{zhang2024deflow} by more than  20\% (\figref{fig:pedestrian}), a $1.5\times$ improvement. Similarly, other state-of-the-art methods like NSFP~\cite{nsfp} and ZeroFlow XL 5x~\cite{vedder2024zeroflow} describe less than 30\% and 20\% of pedestrian motion, respectively.

\oureval{} allows practitioners to effectively compare performance between methods that were almost indistinguishable under Threeway EPE. For example, if you only care about flow performance on cars, DeFlow out-performs all other methods including \ourmethod{} (\figref{fig:car}), while ZeroFlow XL 5x out-performs all other methods on larger vehicles (\figref{fig:other-vehicles}). 

\subsection{What Makes a Good Detector for \ourmethod{}?}\sectionlabel{detectorquality}

As discussed in \sectionref{method}, we tune \emph{\ourframework{}} with a low confidence threshold to maximize recall. What makes a good detector for \ourmethod{}?


We ablate the impact of detector quality on  \ourmethod{} by replacing LE3DE2E ~\cite{wang2023le3de2e} with BEVFusion~\cite{liu2022bevfusion}. We call this new approach \emph{\ourmethodbevfusion{}}. BEVFusion only has 2\% lower mAP than LE3DE2E on the AV2 detection leaderboard\footnote{BEVFusion~\cite{liu2022bevfusion} was second on the \emph{Argoverse 2 2023 3D Detection, Tracking and Forecasting challenge}~\cite{peri2022towards, peri2023empirical, Peri_2022_CVPR}.}, but we find that \ourmethodbevfusion{} performs significantly worse than \ourmethod{}, with 10\% to 22\% drops in performance on Dynamic Normalized EPE (\tableref{tab:bevfusionresults}).

This significant degradation is the result of BEVFusion's poor recall at low confidence thresholds (\tableref{tab:bevfusionconf}). In contrast, LE3DE2E has very high recall at low thresholds (\figref{fig:bevfusionrecall}), producing many candidate boxes for pedestrians in the scene. BEVFusion's false negatives are extremely costly to \ourmethodbevfusion{}, as they result in $\zerovec$ flow estimates that miss 100\% of each false positive pedestrian's motion.

More broadly, a good detector for \ourframework{} isn't necessarily one with a high mAP; it's is one with very high recall and accurate heading estimates. Notably, these error characteristics enable the tracker to reject false positives. We believe this interaction between the detector and tracker is an important yet subtle point --- two detectors may have the same mAP, but dramatically different performance in our \ourframework{} framework.

\begin{figure}[htbp]
\centering
\begin{subfigure}{.16\textwidth}
  \centering
  \includegraphics[width=\linewidth]{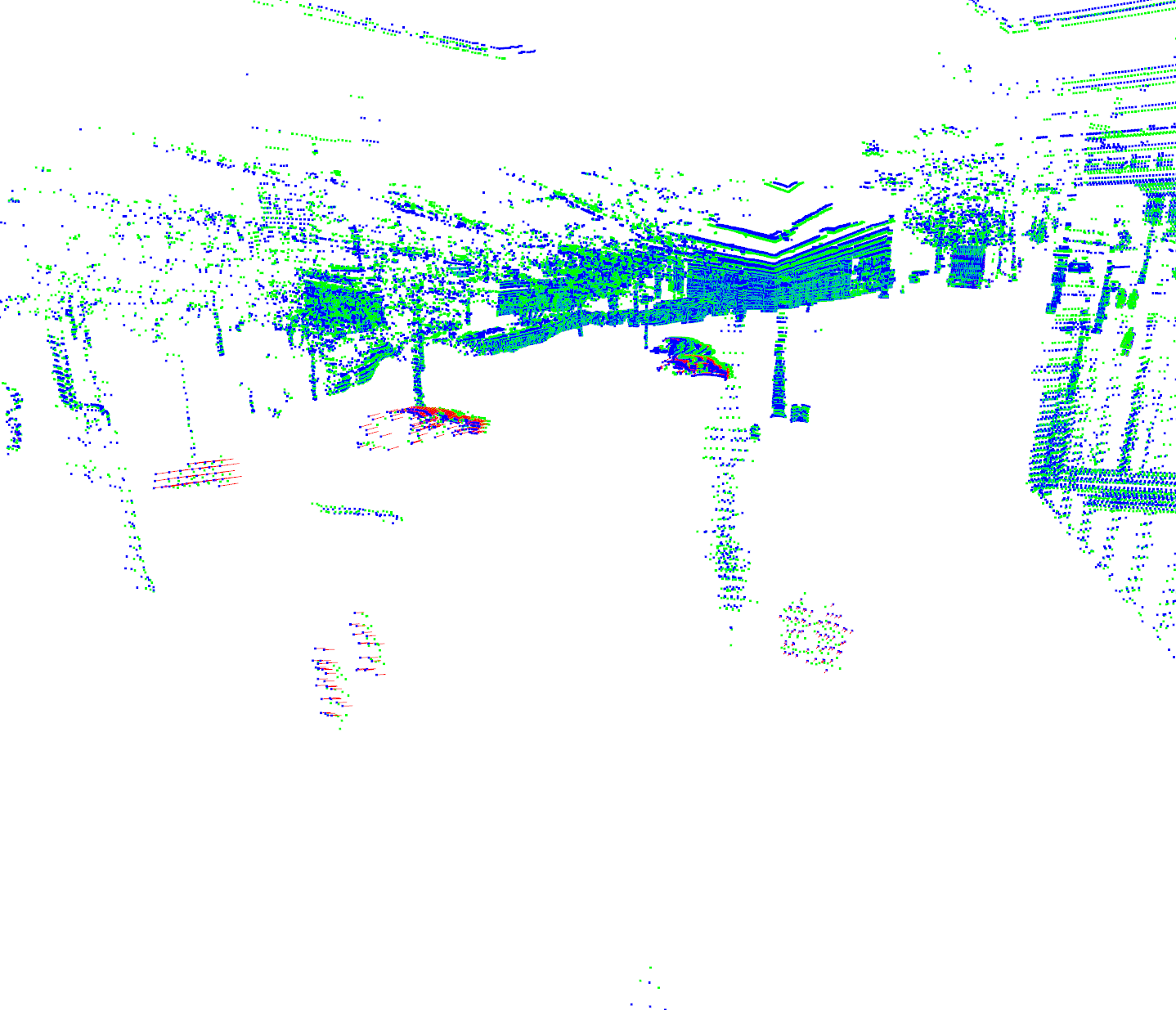}
  \includegraphics[width=\linewidth]{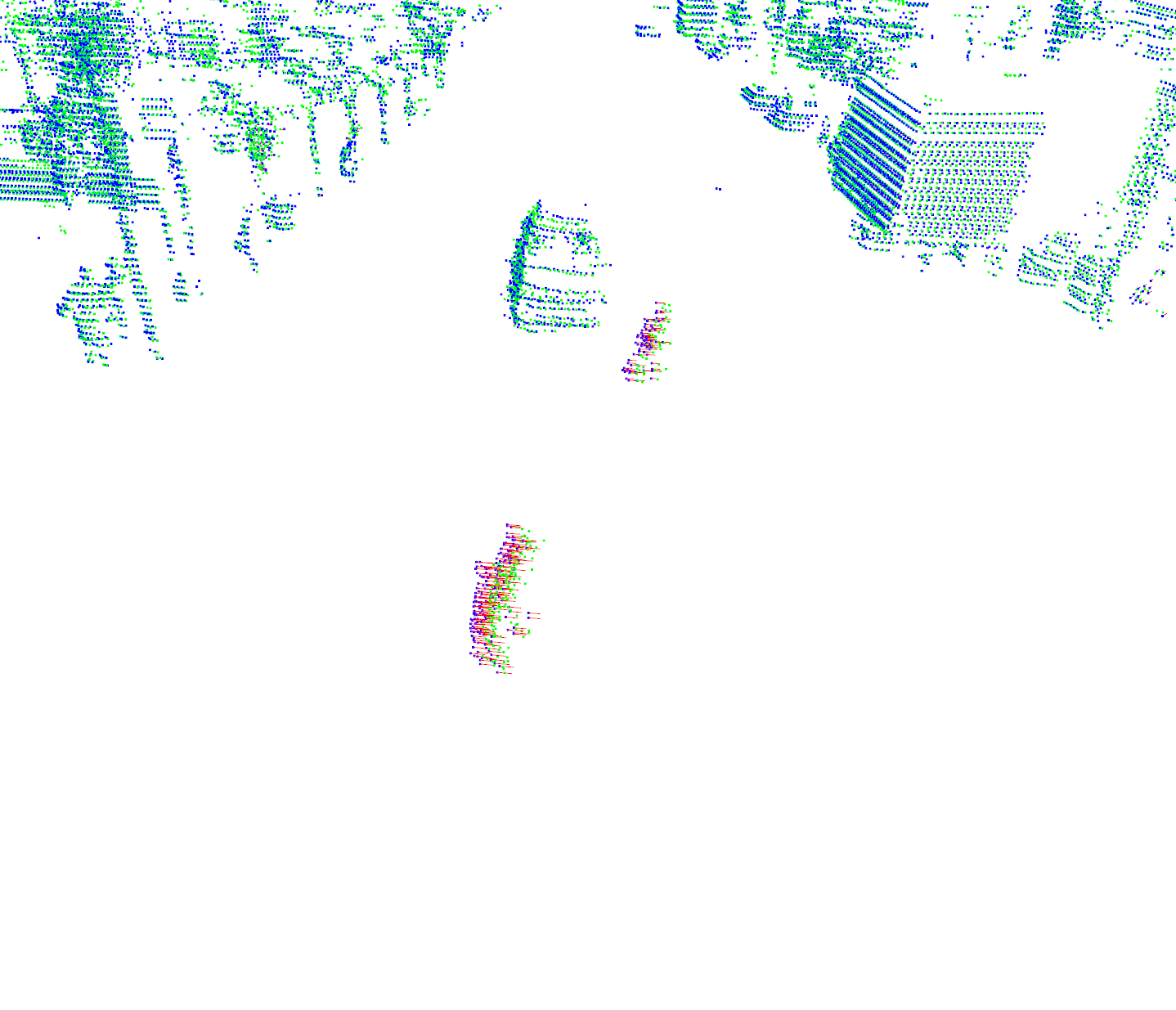}
  \includegraphics[width=\linewidth]{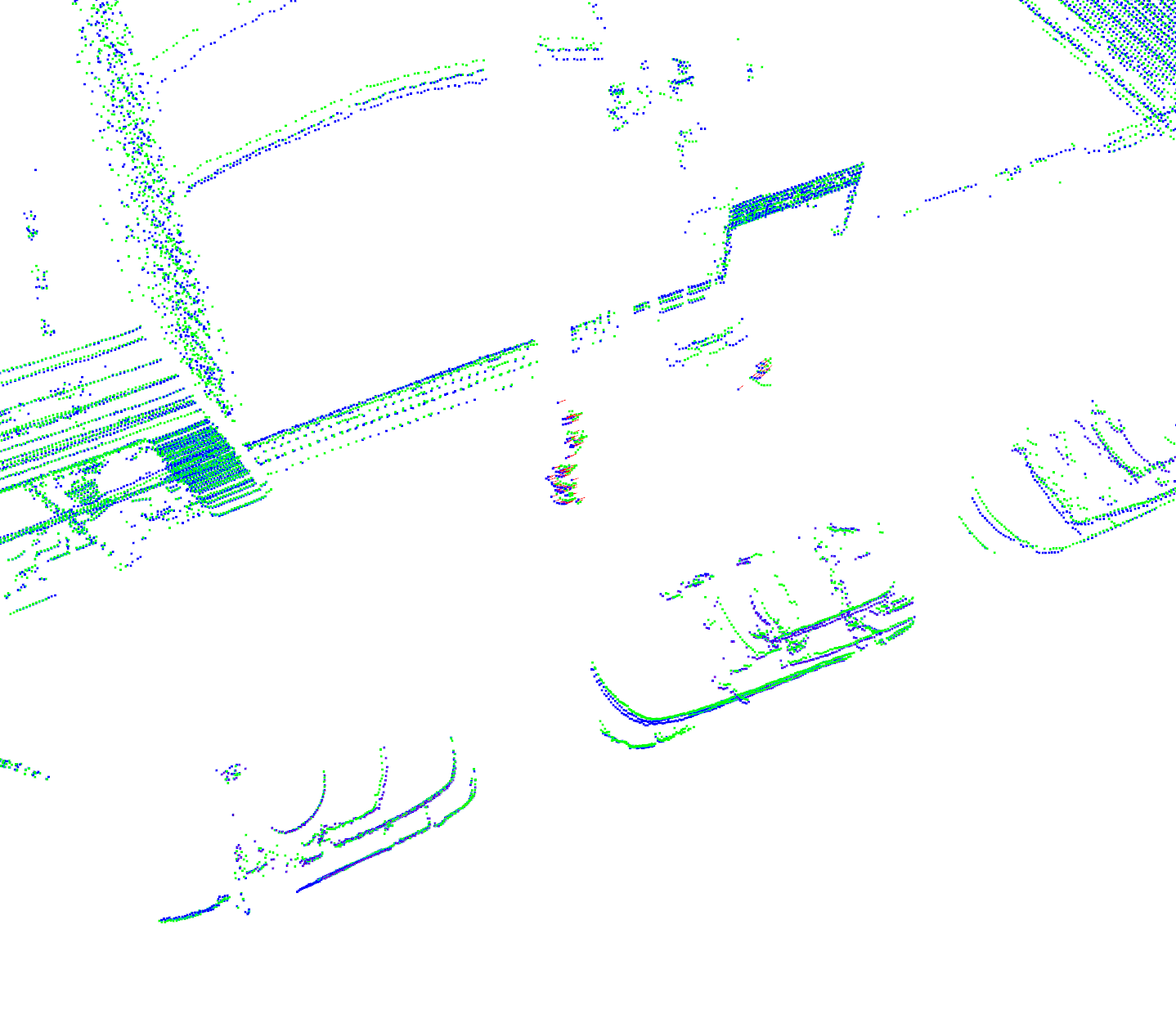}
  \includegraphics[width=\linewidth]{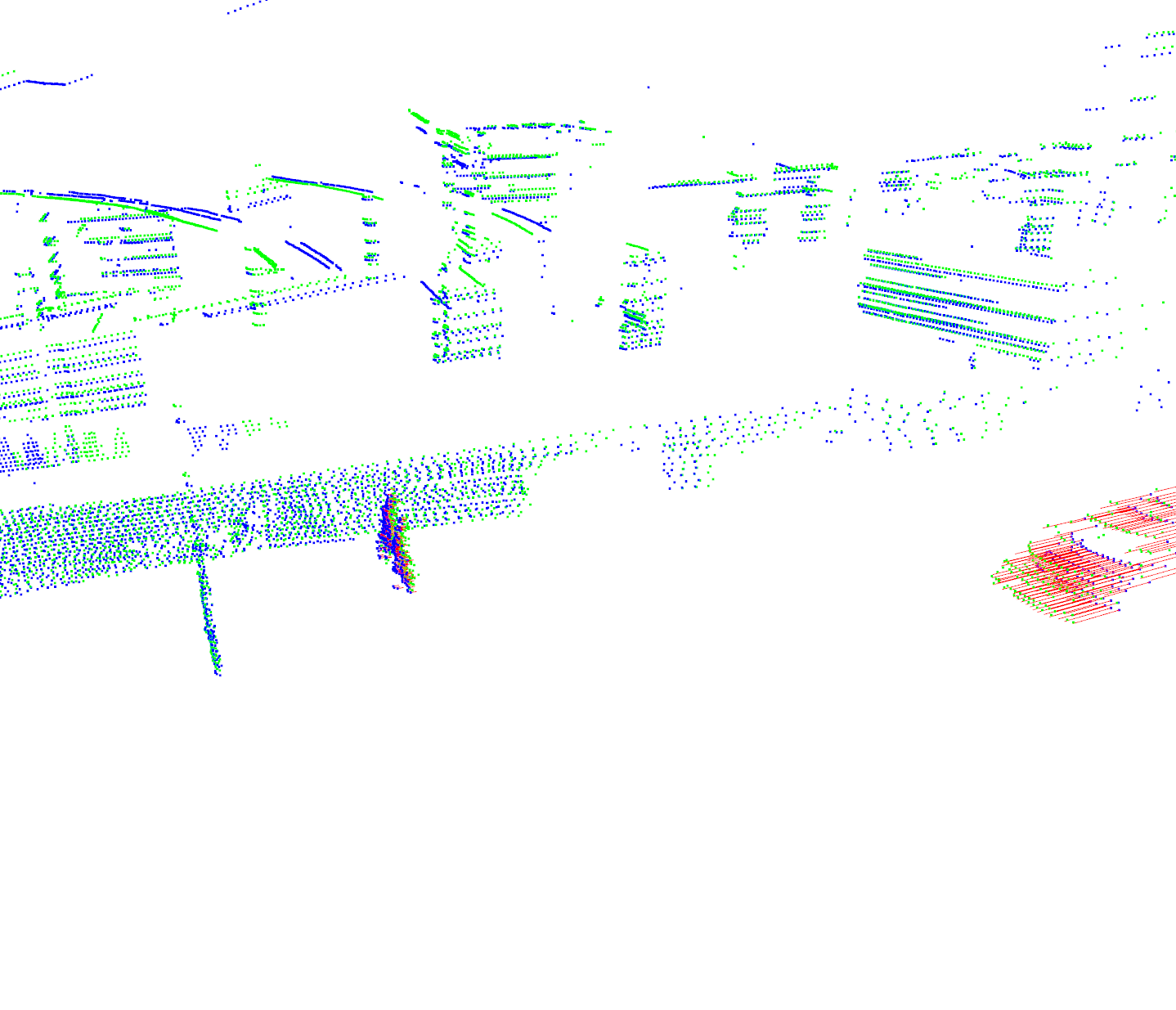}
  \includegraphics[width=\linewidth]{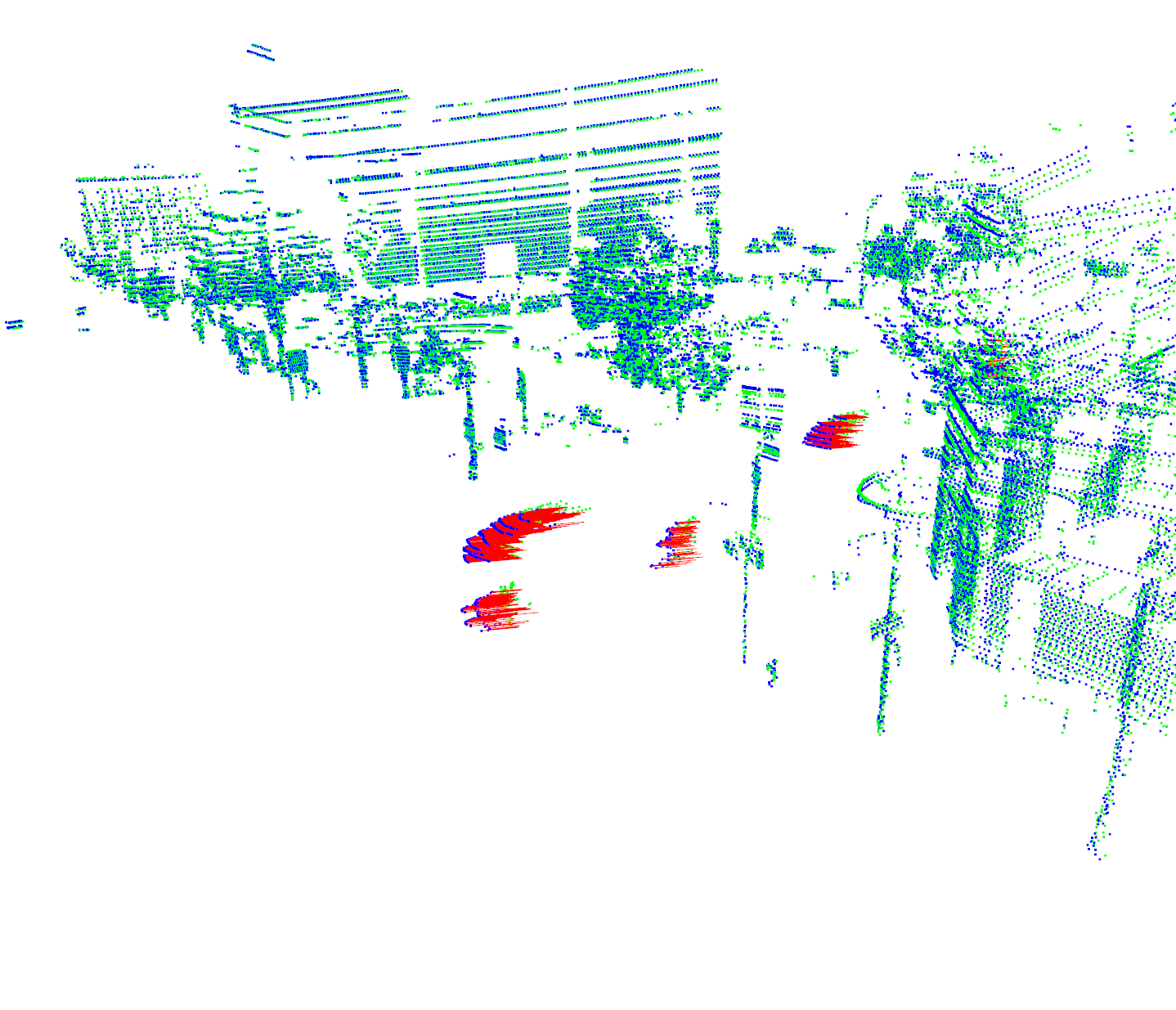}
  \includegraphics[width=\linewidth]{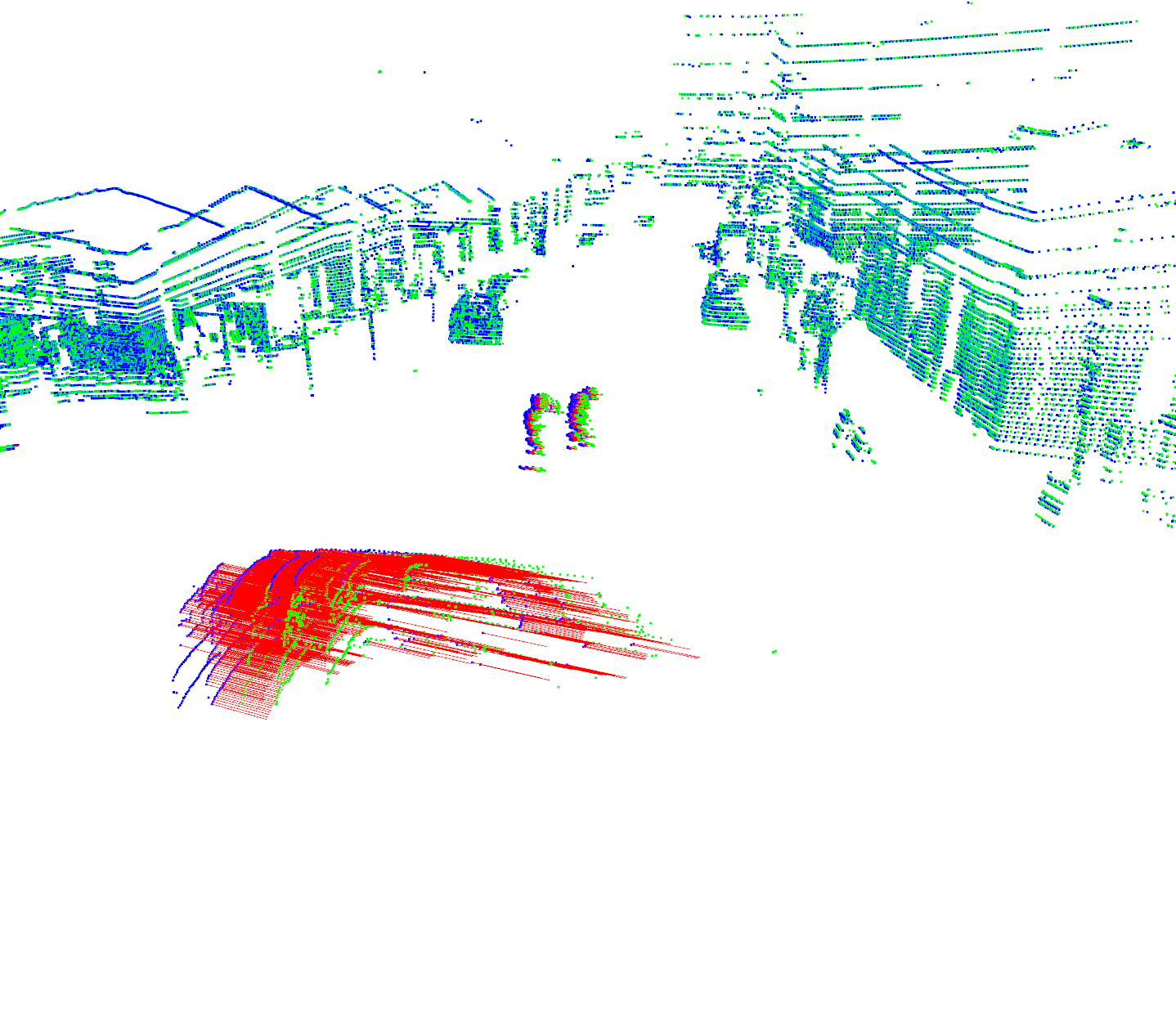}
  \caption{\tinier{Ground Truth}}
  \label{fig:sub5}
\end{subfigure}%
\begin{subfigure}{.16\textwidth}
  \centering
  \includegraphics[width=\linewidth]{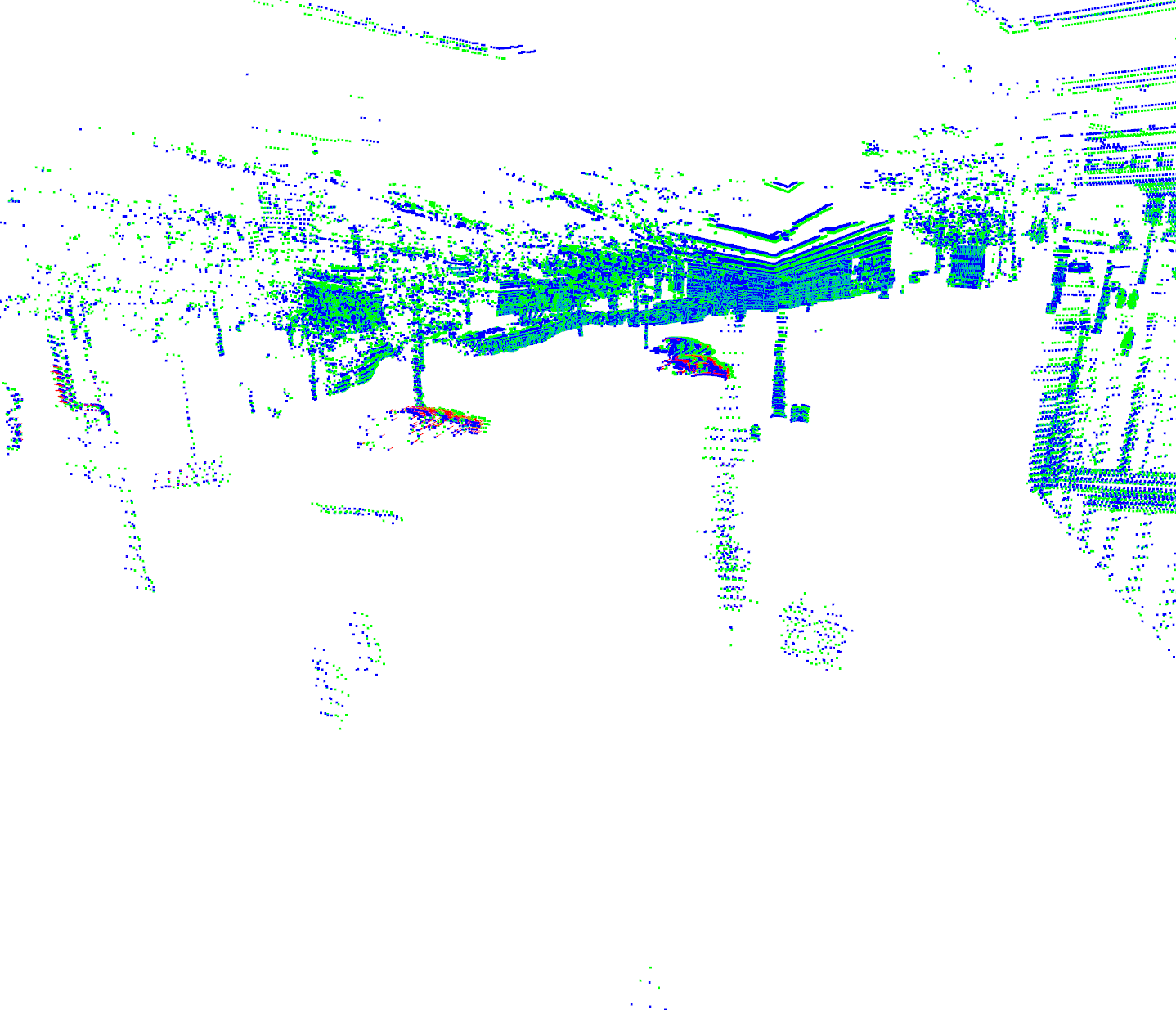}
  \includegraphics[width=\linewidth]{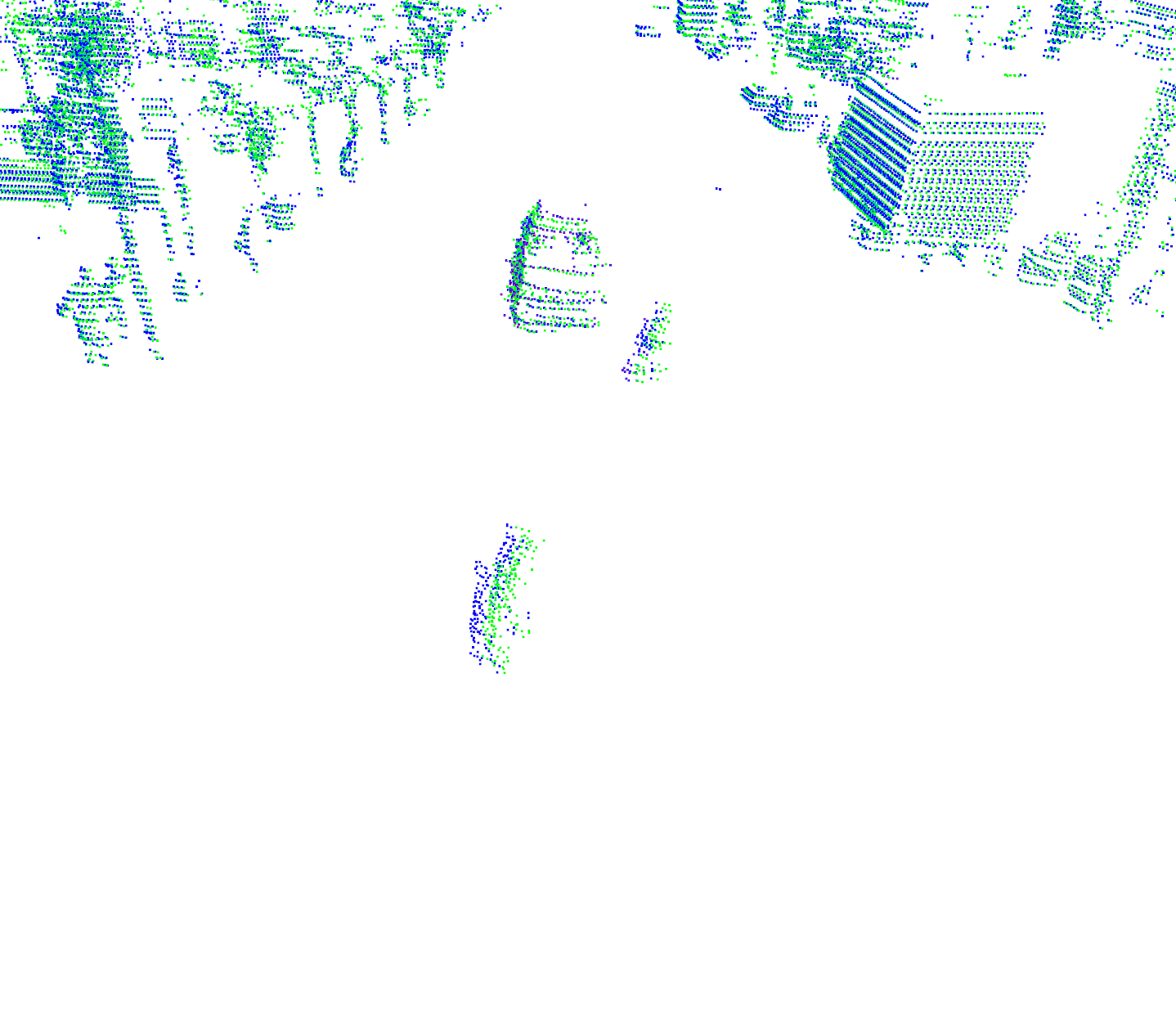}
  \includegraphics[width=\linewidth]{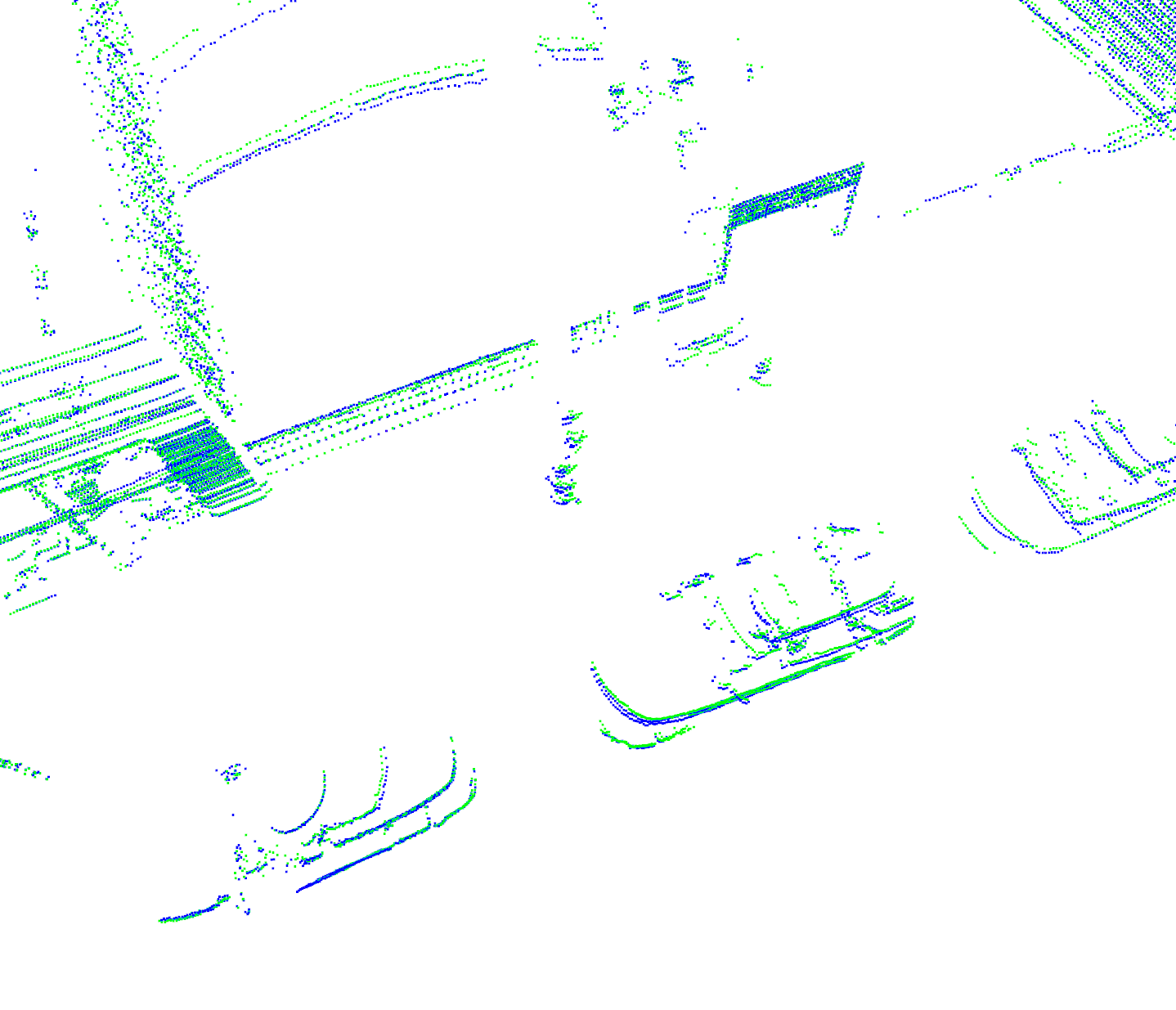}
  \includegraphics[width=\linewidth]{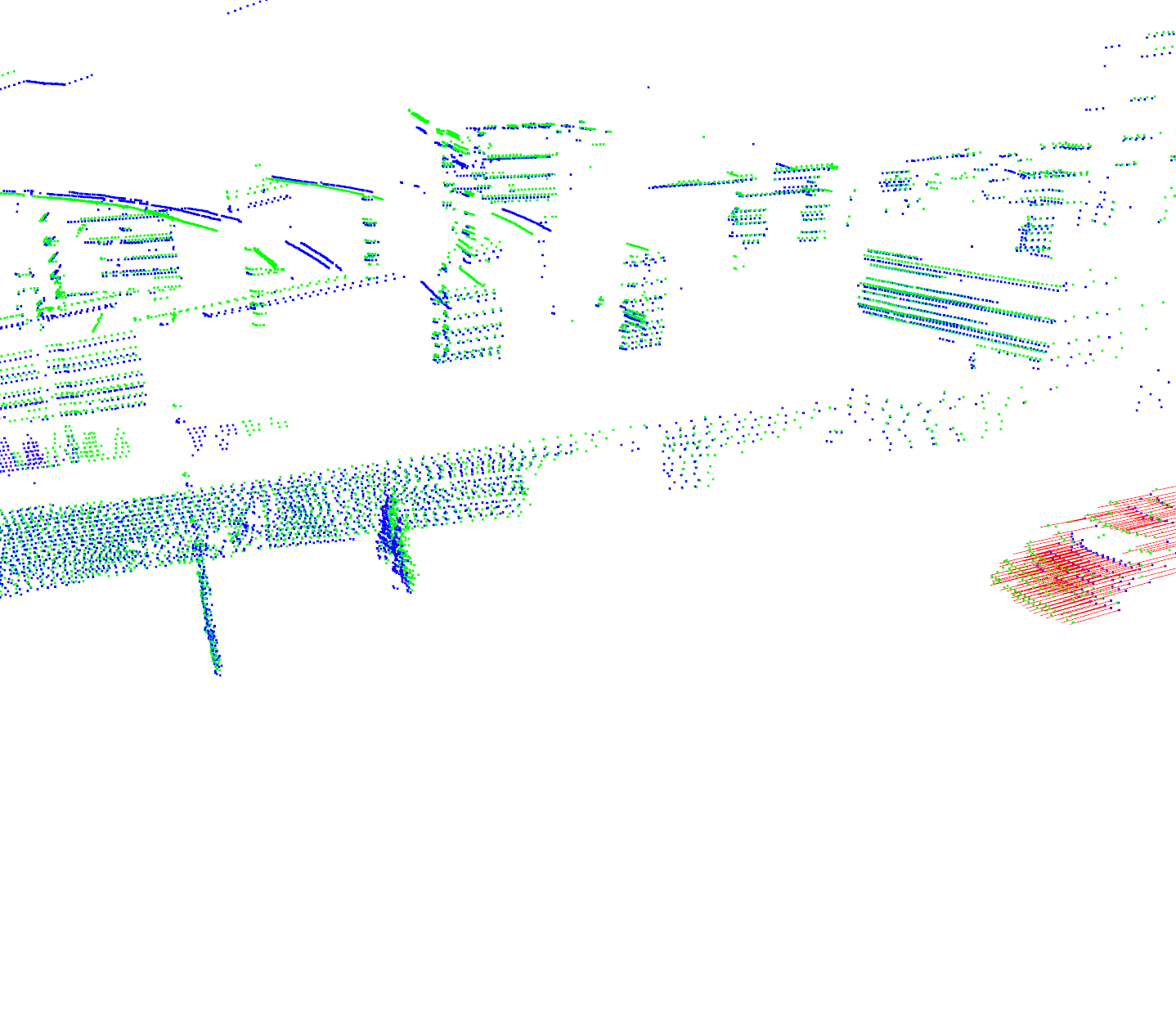}
  \includegraphics[width=\linewidth]{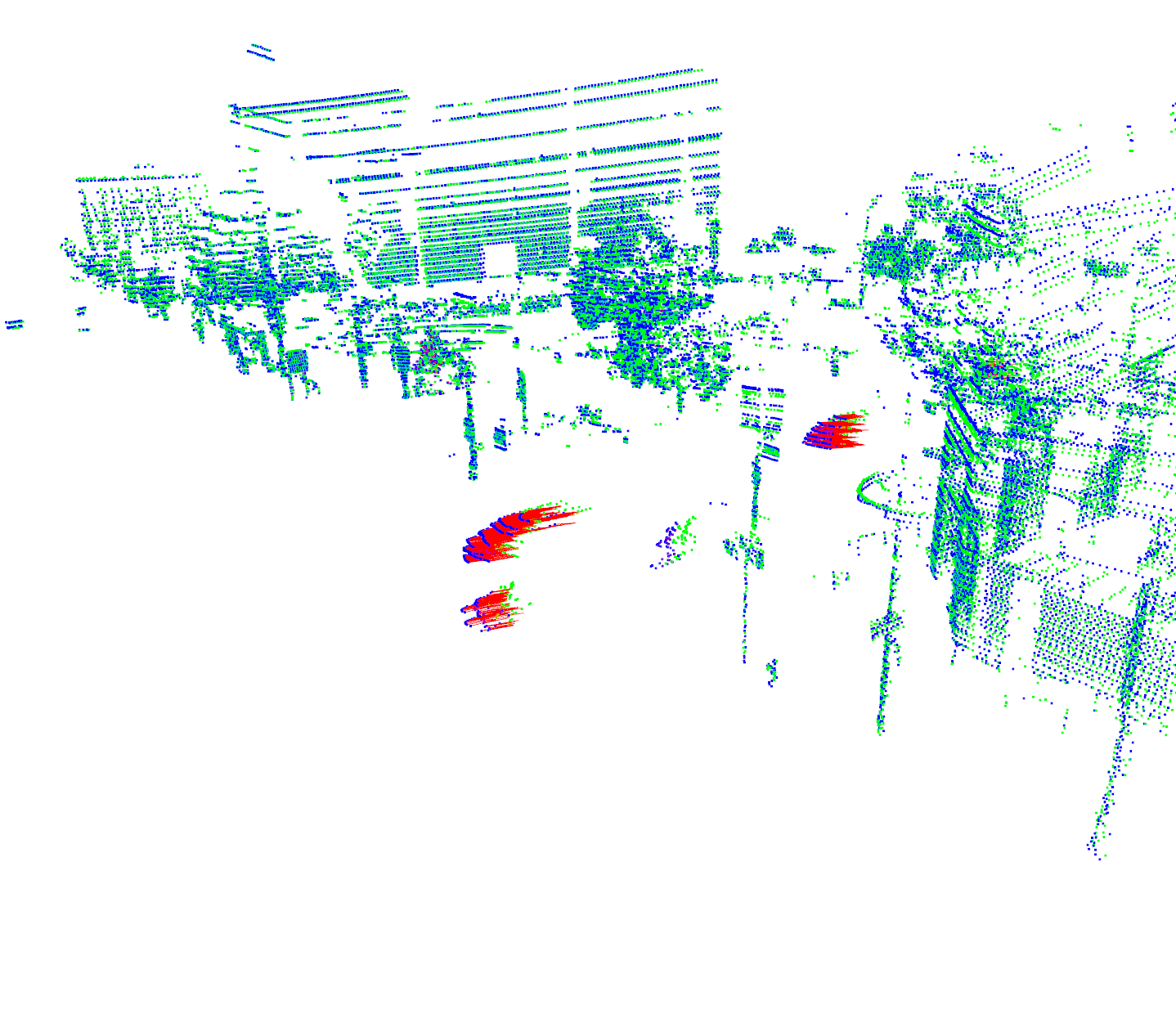}
  \includegraphics[width=\linewidth]{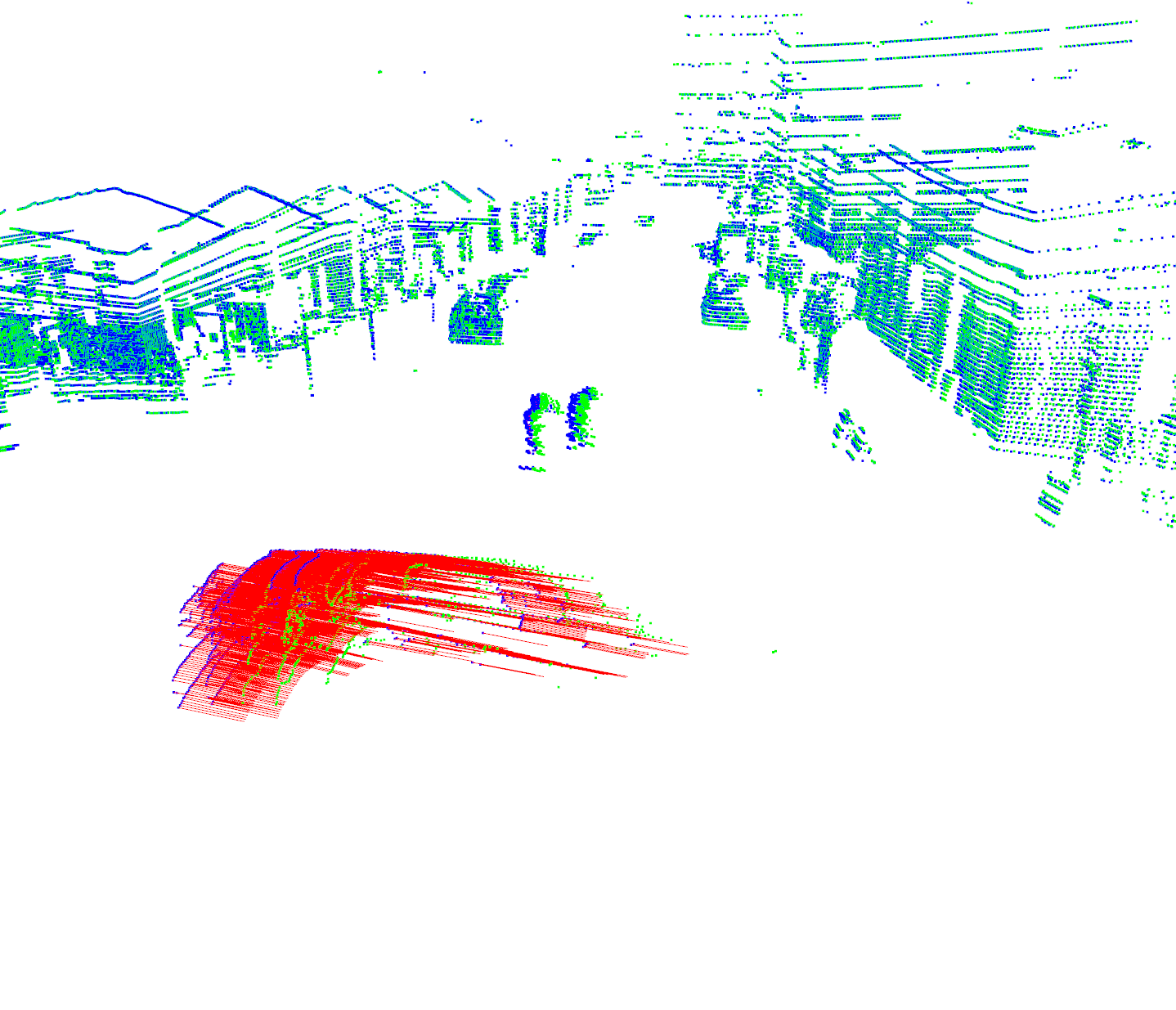}
  \caption{\tinier FastFlow3D}
  \label{fig:sub2}
\end{subfigure}
\begin{subfigure}{.16\textwidth}
  \centering
  \includegraphics[width=\linewidth]{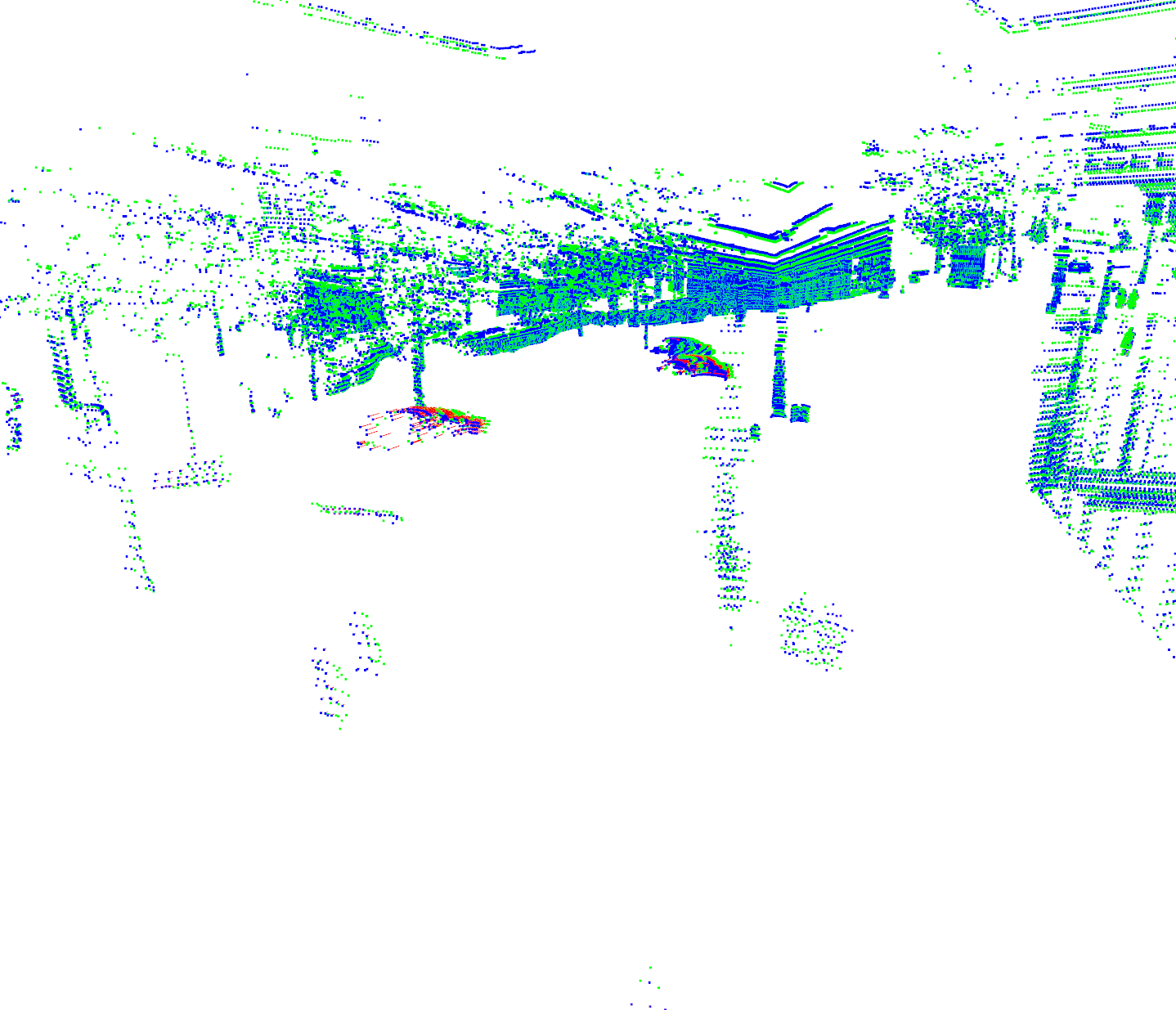}
  \includegraphics[width=\linewidth]{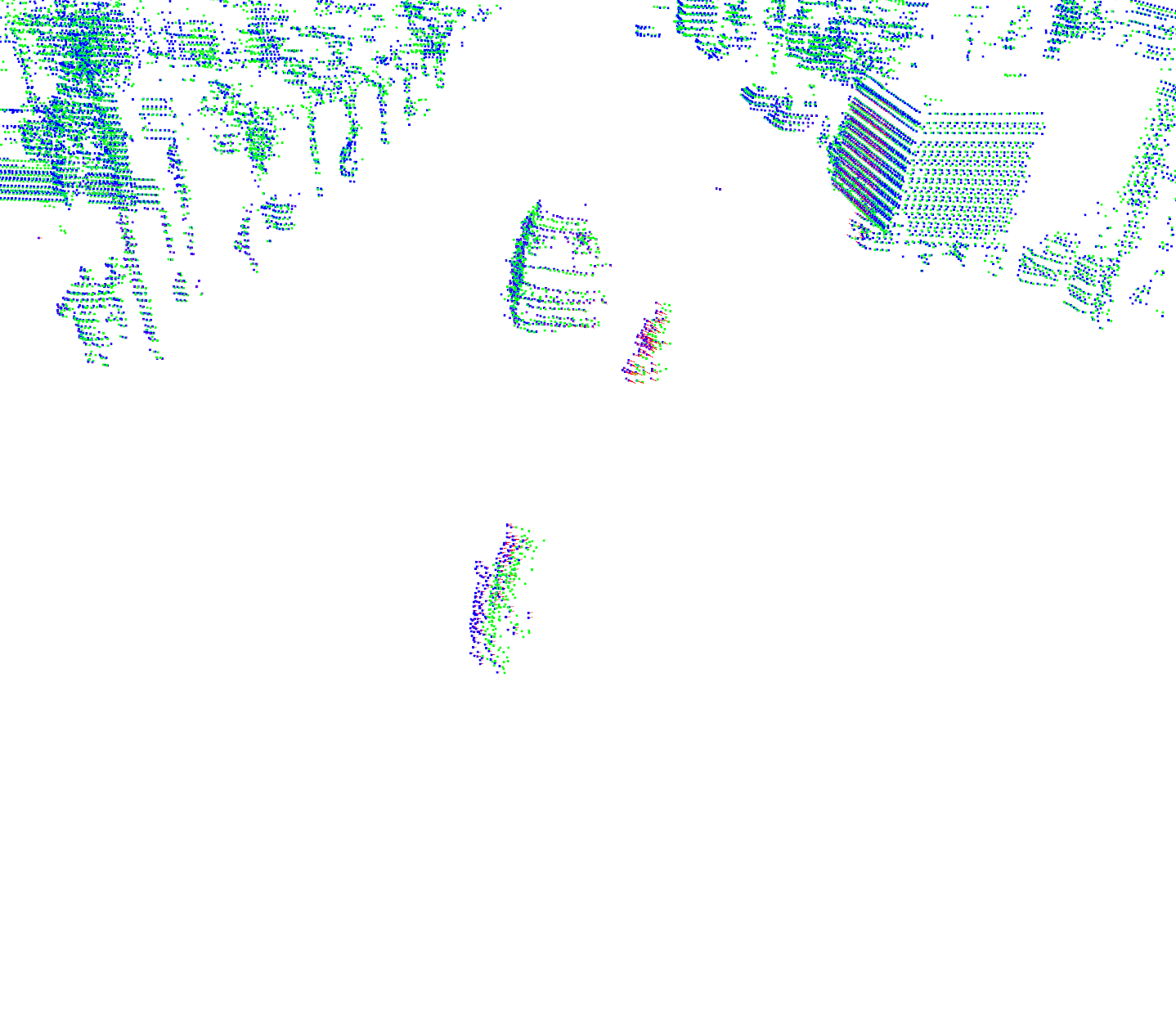}
  \includegraphics[width=\linewidth]{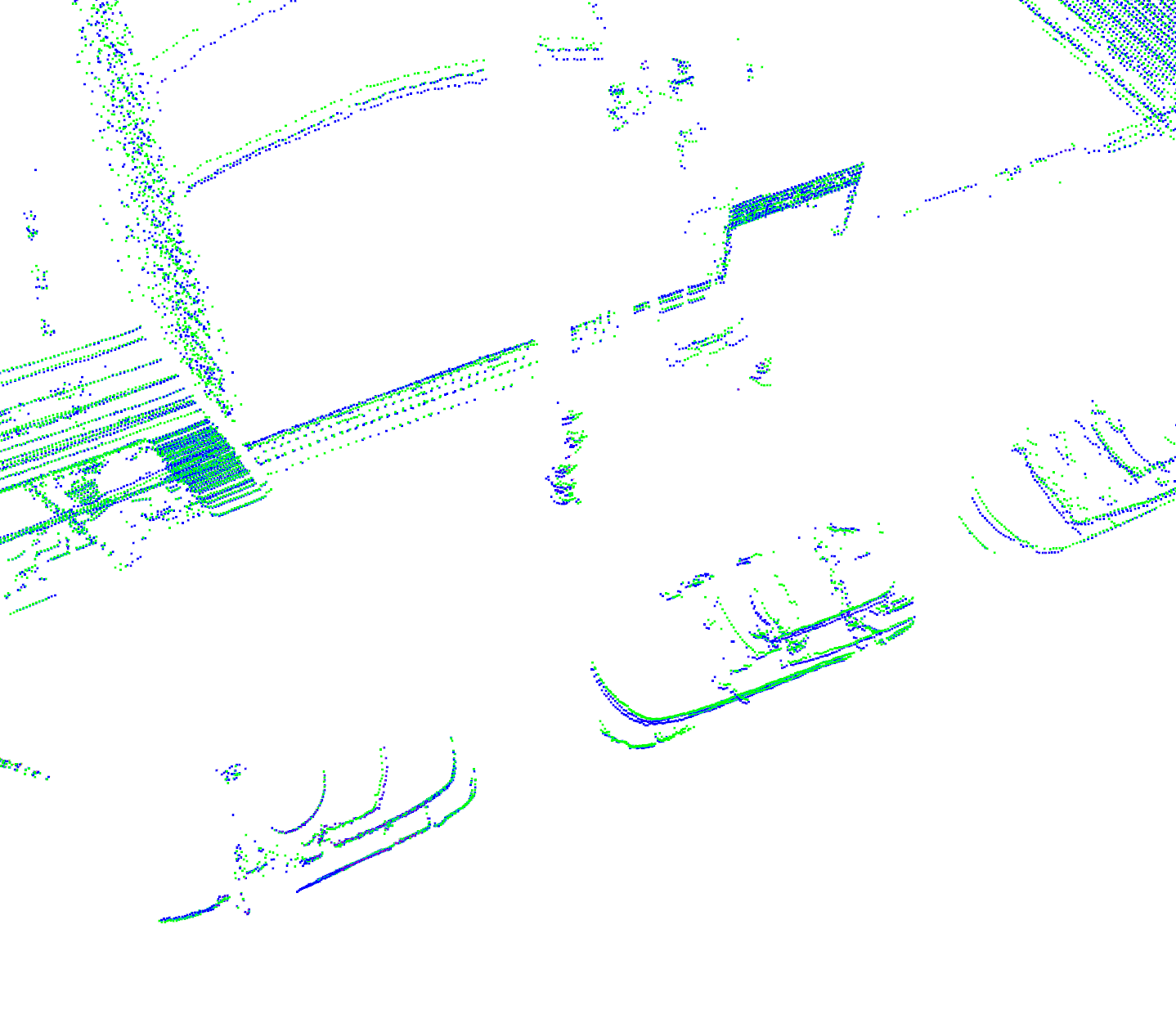}
  \includegraphics[width=\linewidth]{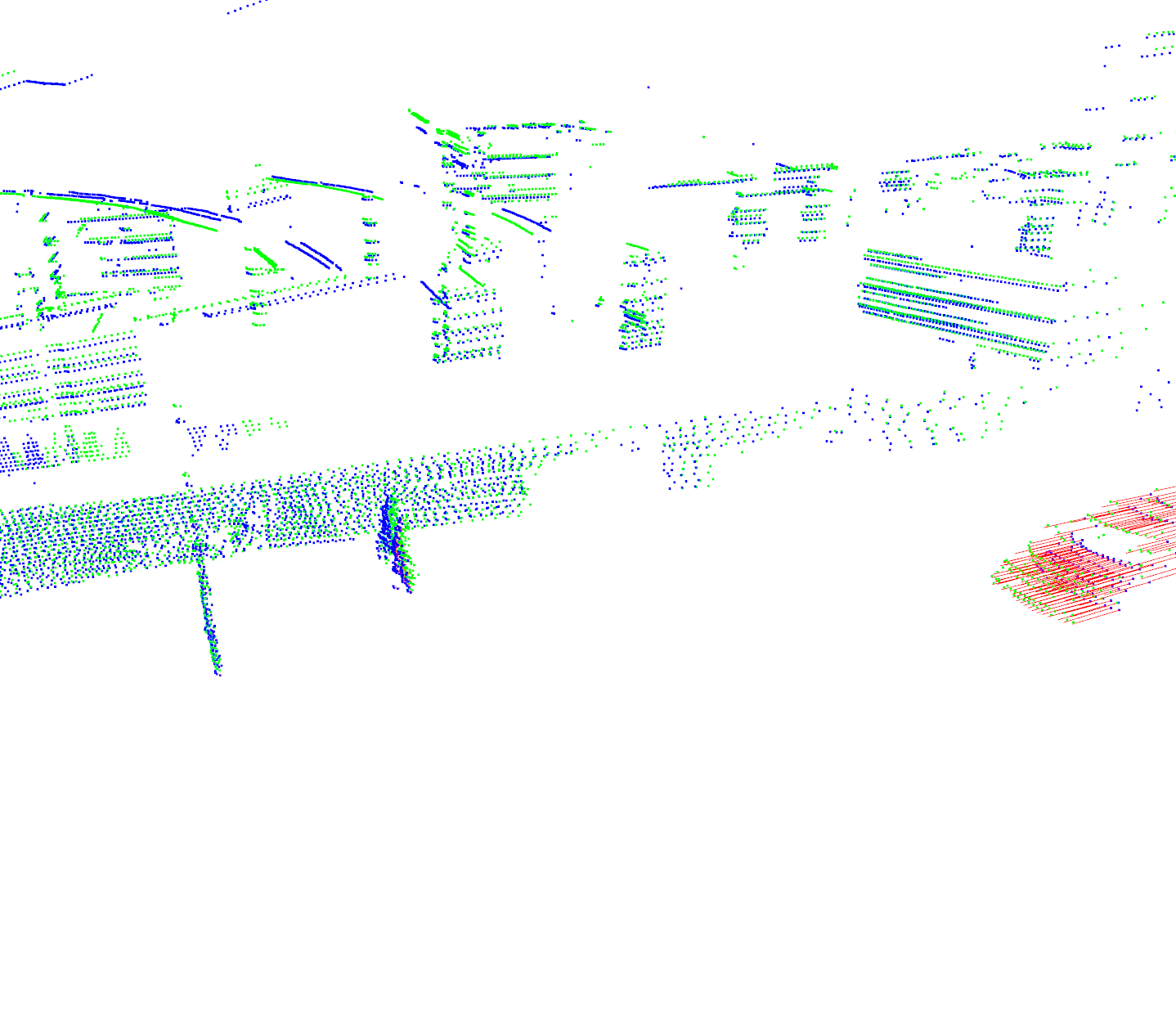}
  \includegraphics[width=\linewidth]{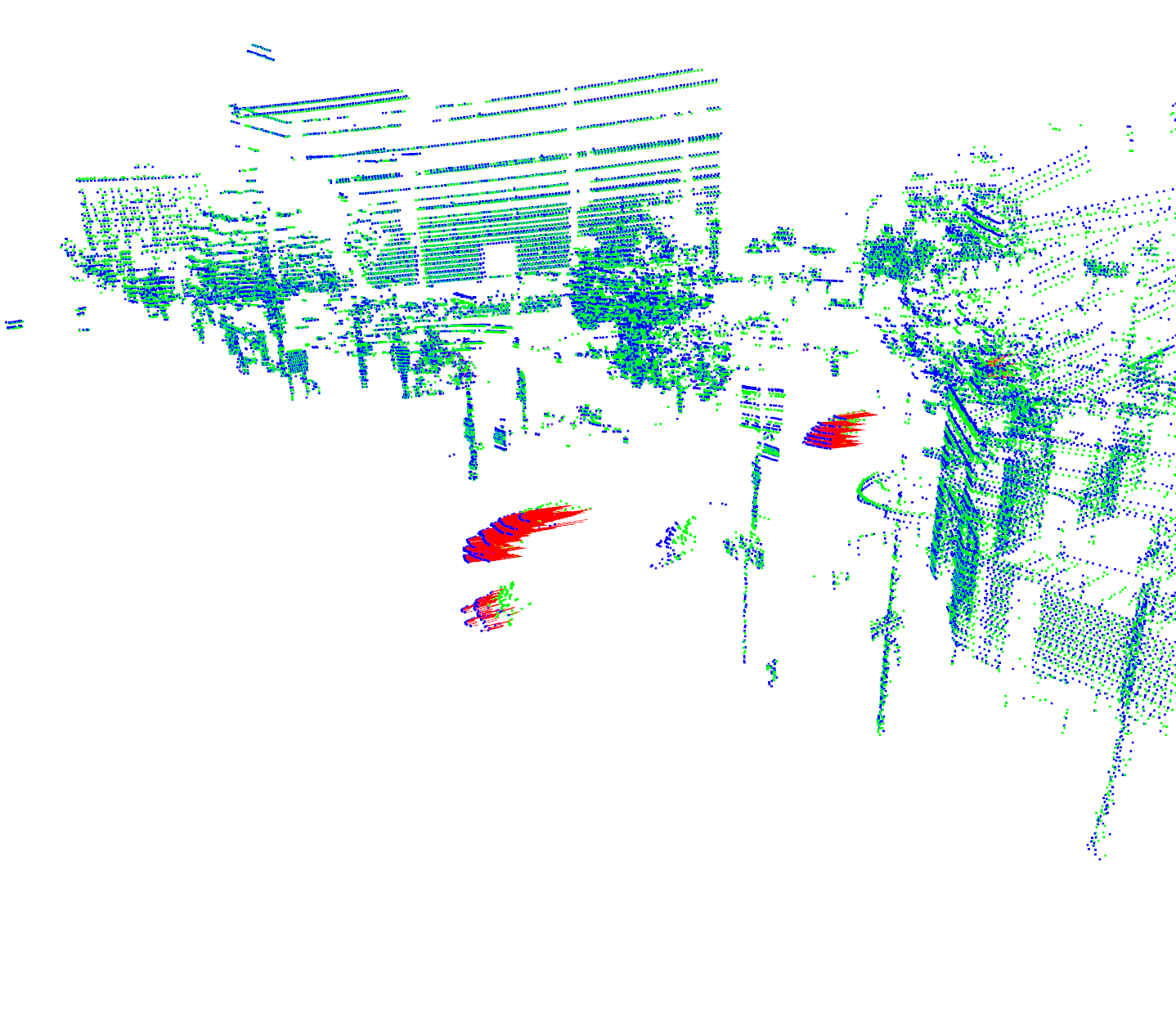}
  \includegraphics[width=\linewidth]{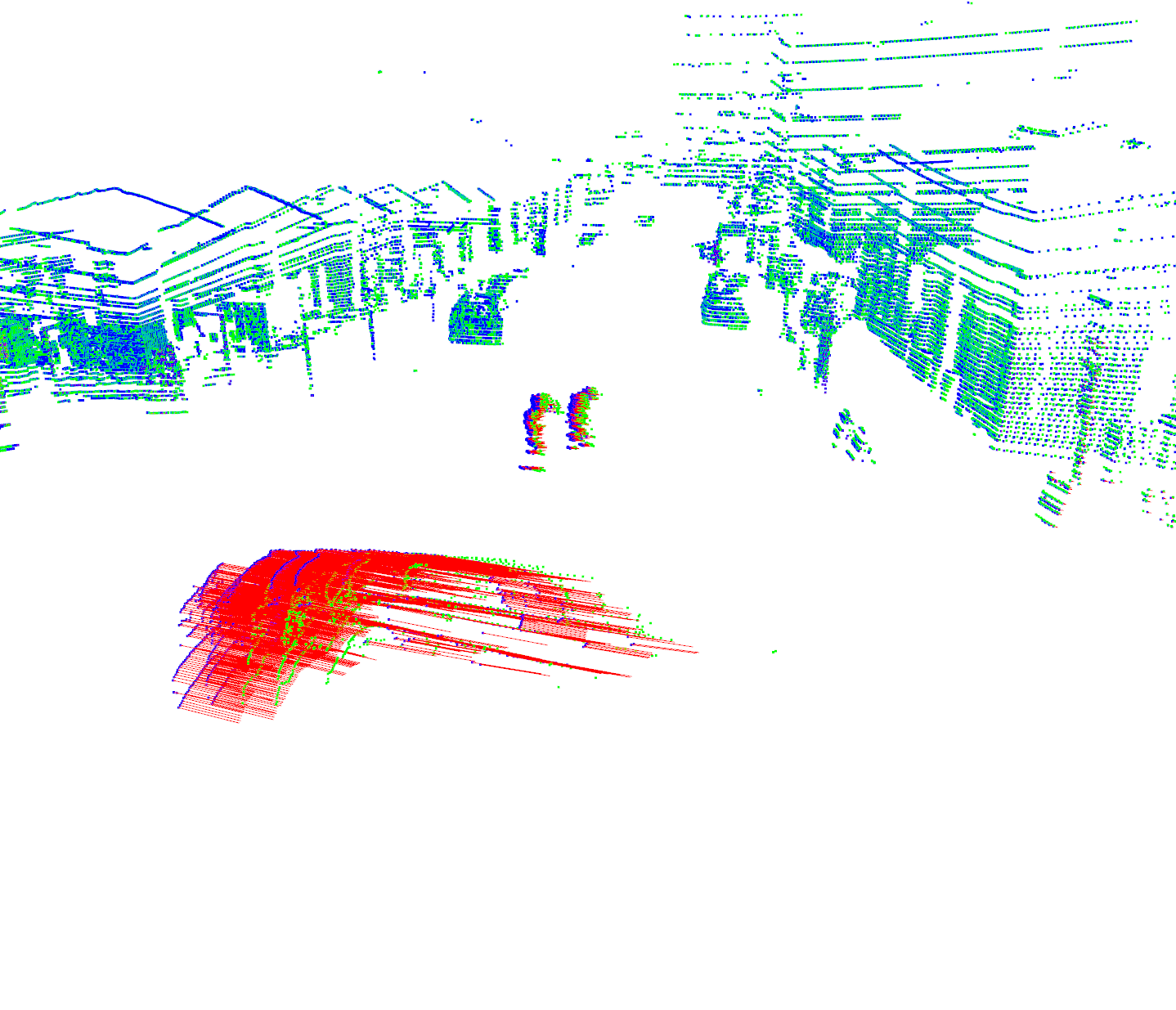}
  \caption{\tinier DeFlow}
  \label{fig:sub3}
\end{subfigure}
\begin{subfigure}{.16\textwidth}
  \centering
  \includegraphics[width=\linewidth]{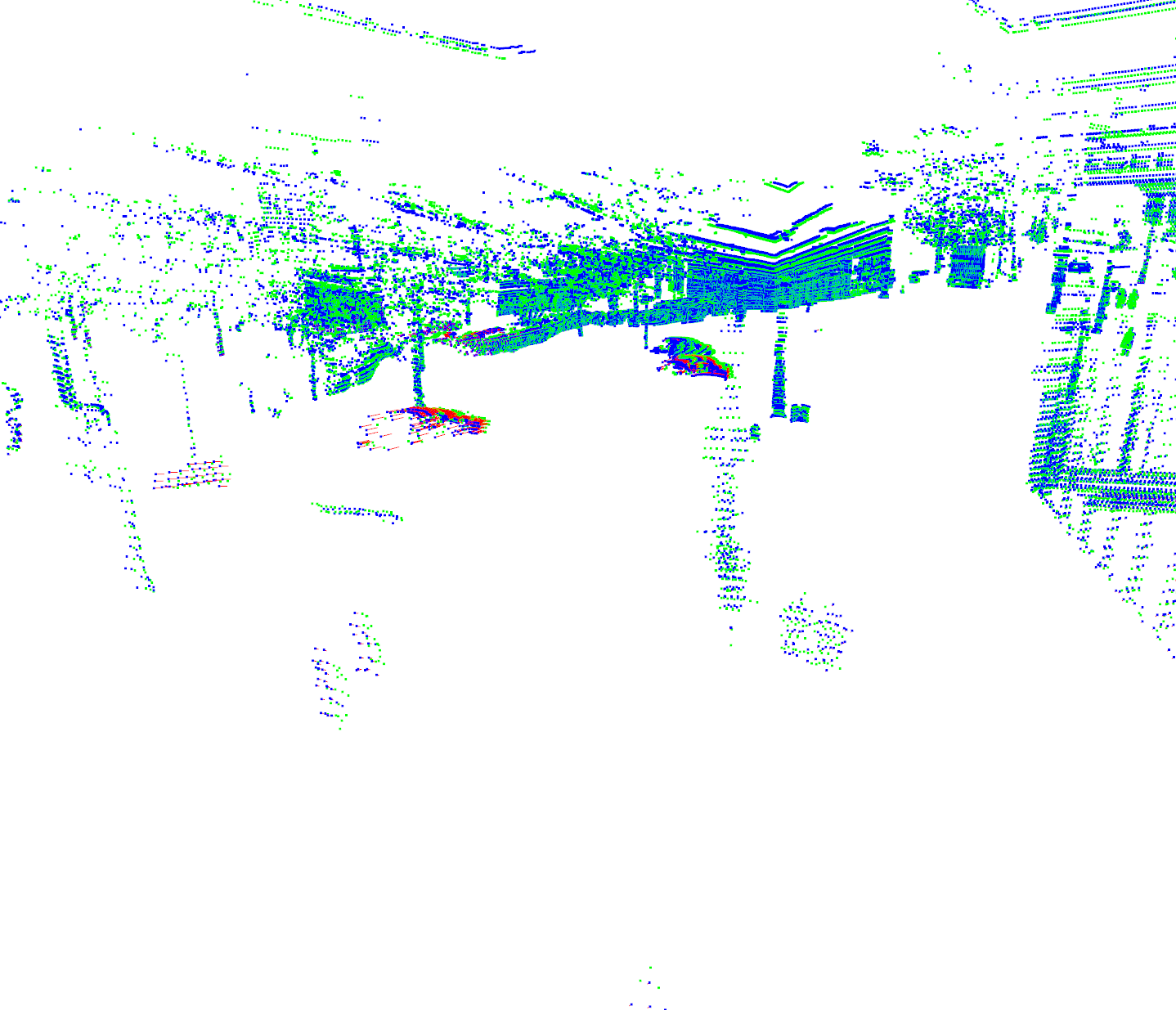}
  \includegraphics[width=\linewidth]{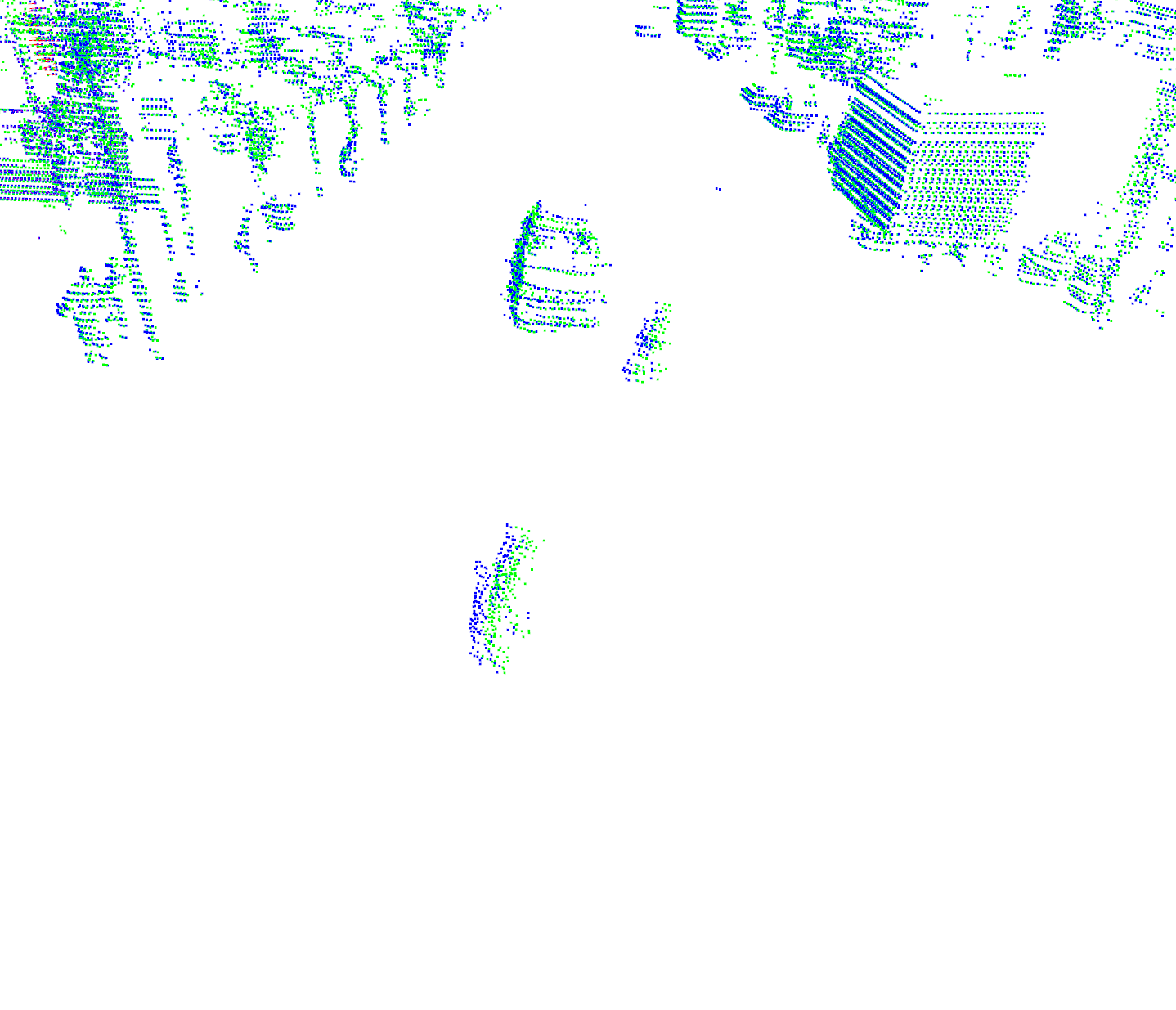}
  \includegraphics[width=\linewidth]{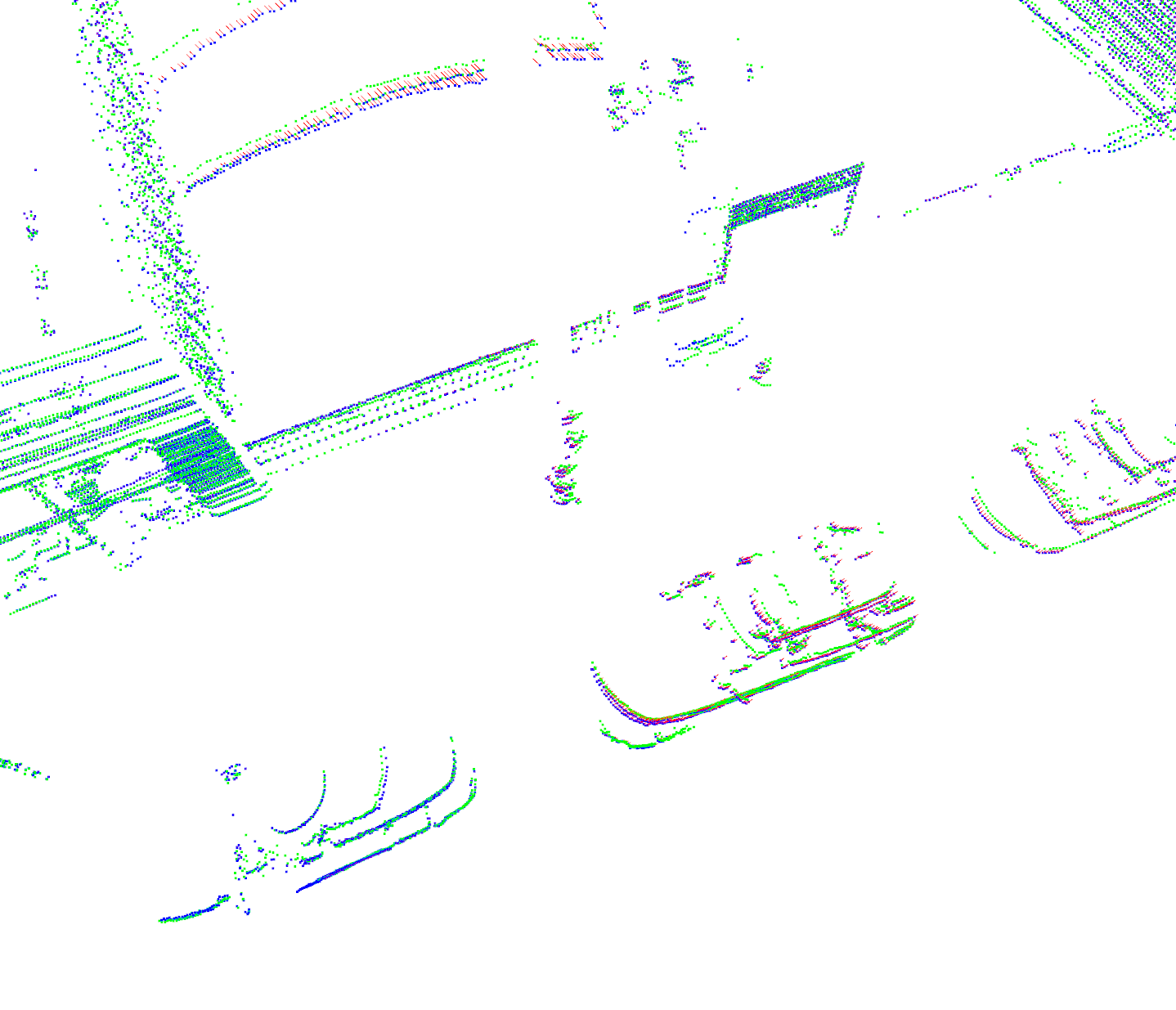}
  \includegraphics[width=\linewidth]{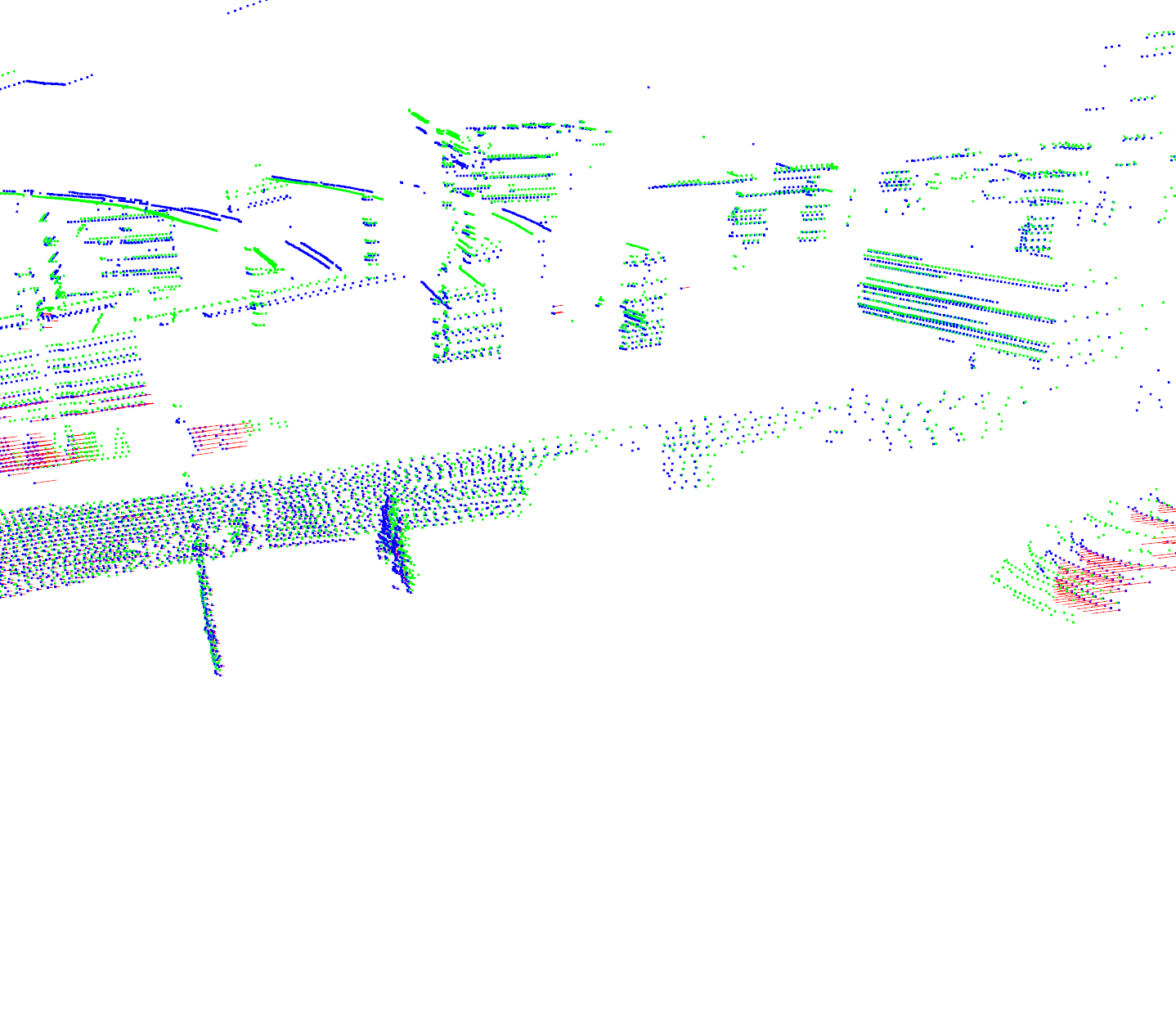}
  \includegraphics[width=\linewidth]{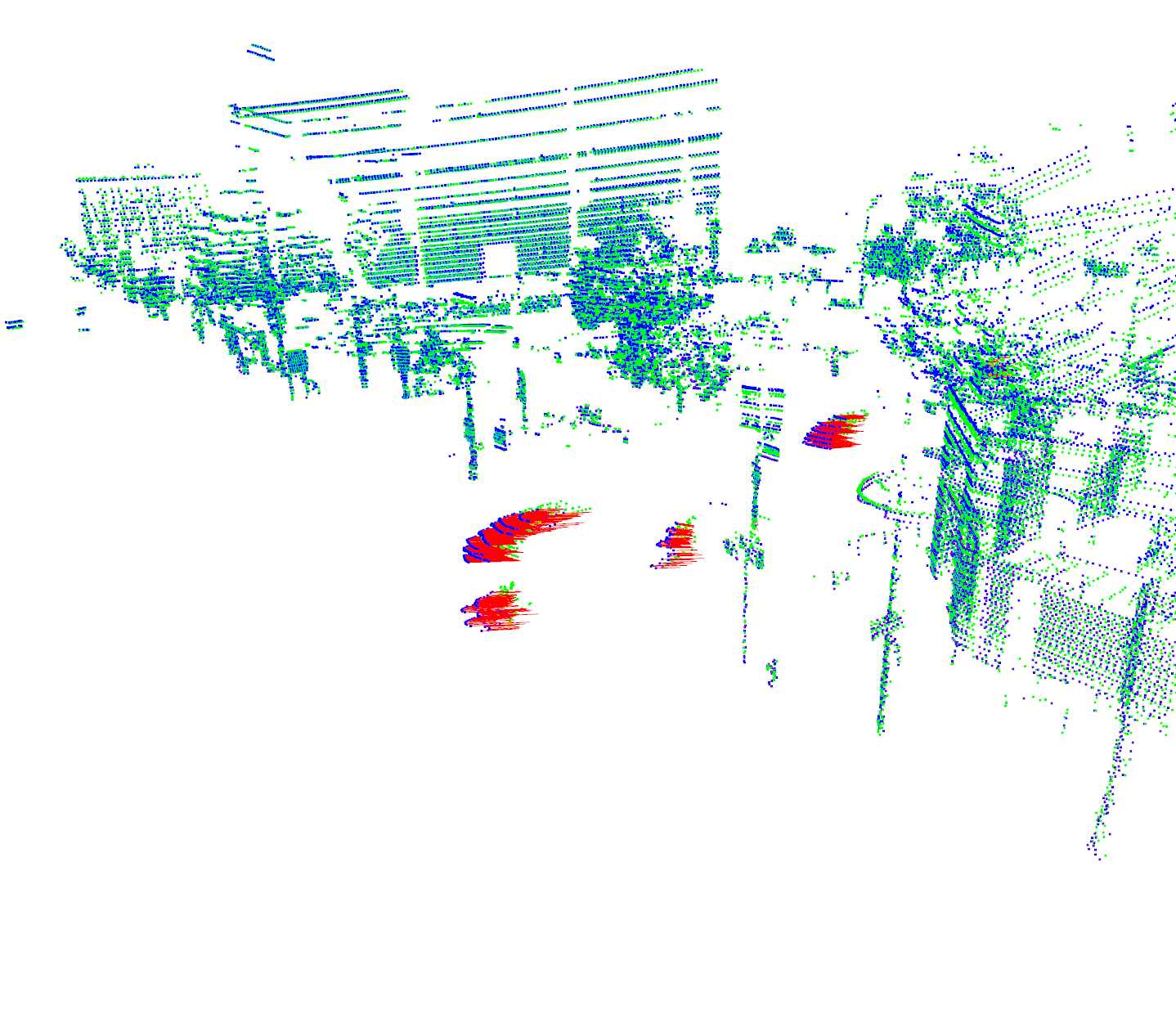}
  \includegraphics[width=\linewidth]{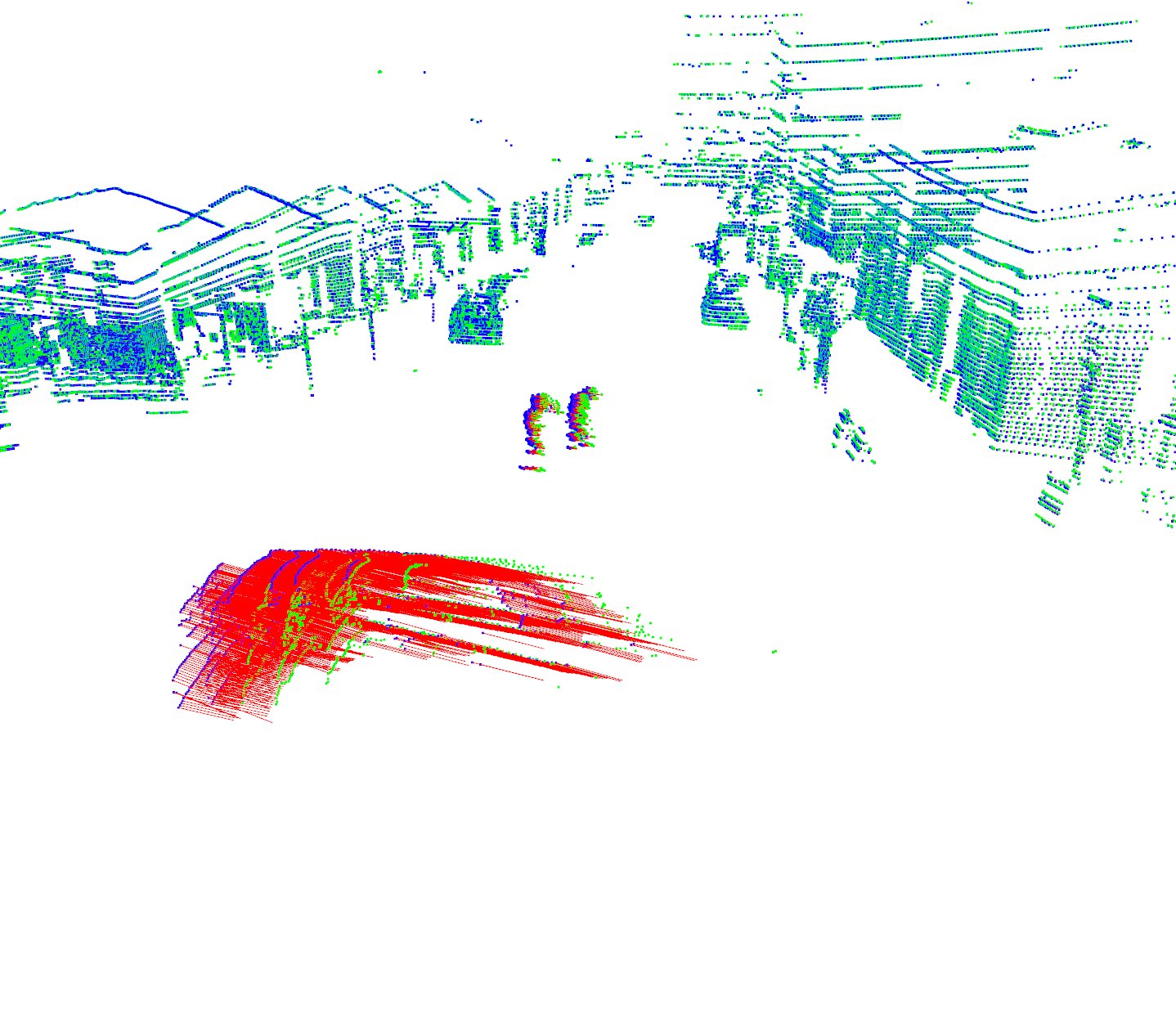}
  \caption{\tinier NSFP}
  \label{fig:sub4}
\end{subfigure}%
\begin{subfigure}{.16\textwidth}
  \centering
  \includegraphics[width=\linewidth]{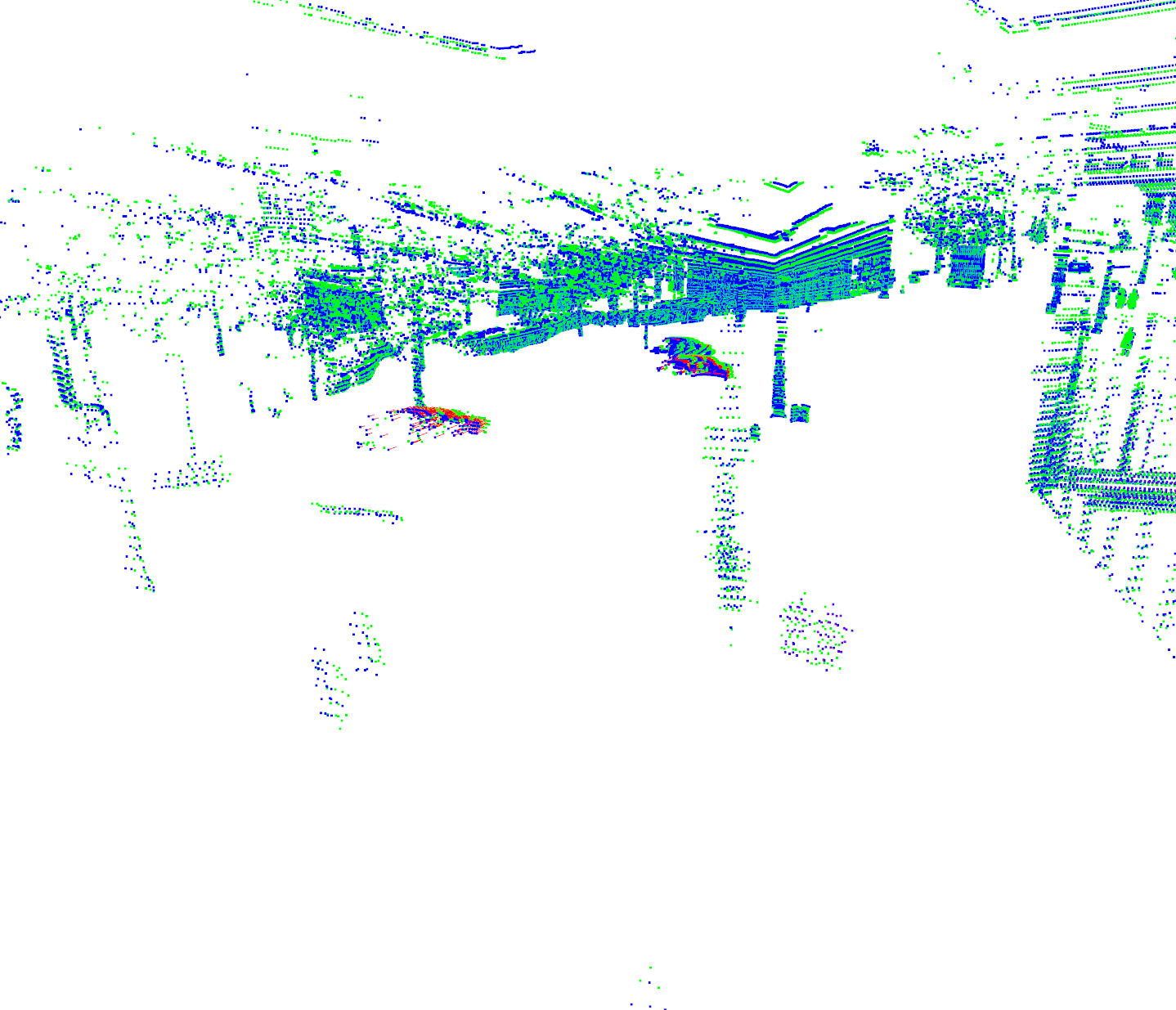}
  \includegraphics[width=\linewidth]{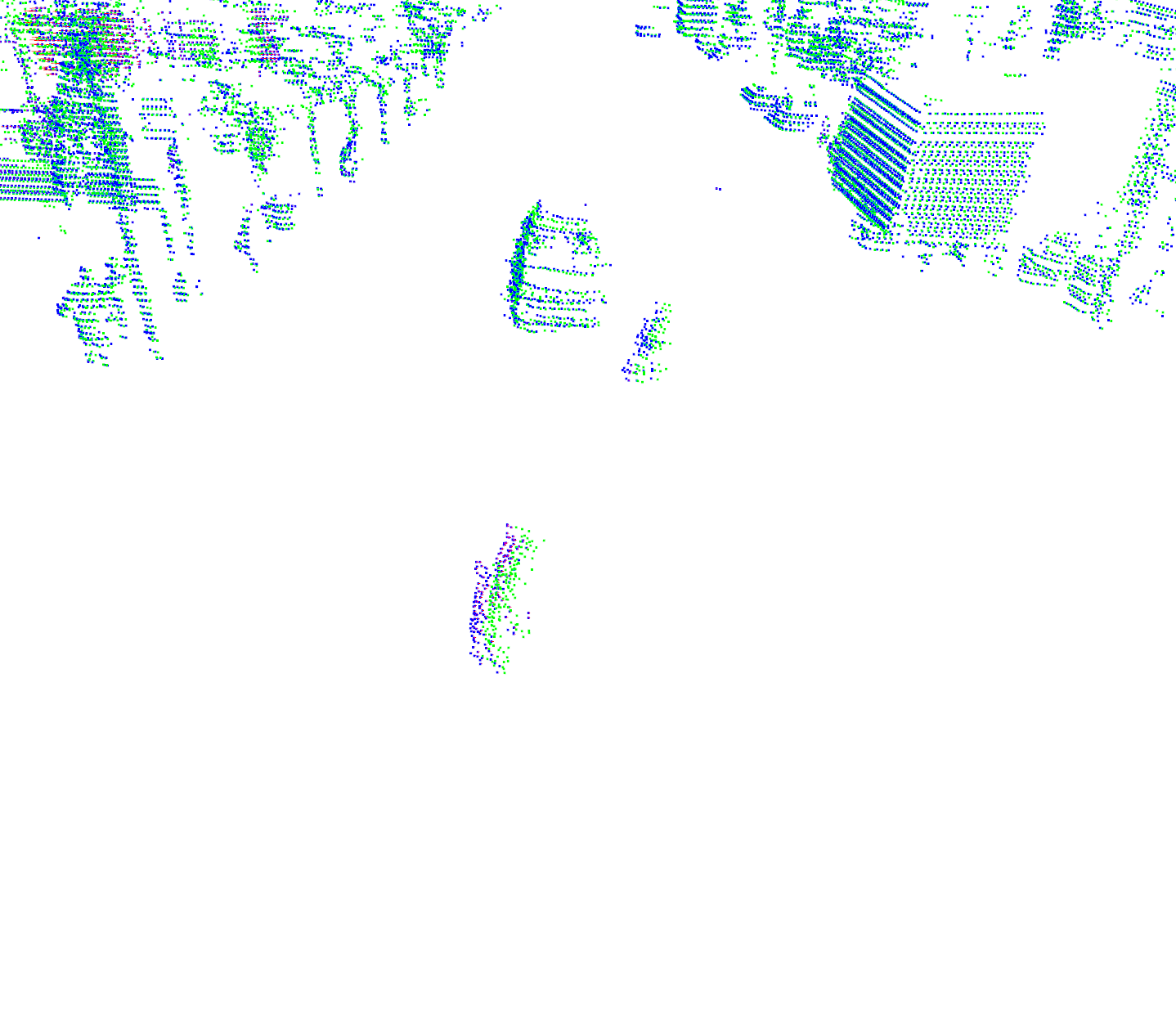}
  \includegraphics[width=\linewidth]{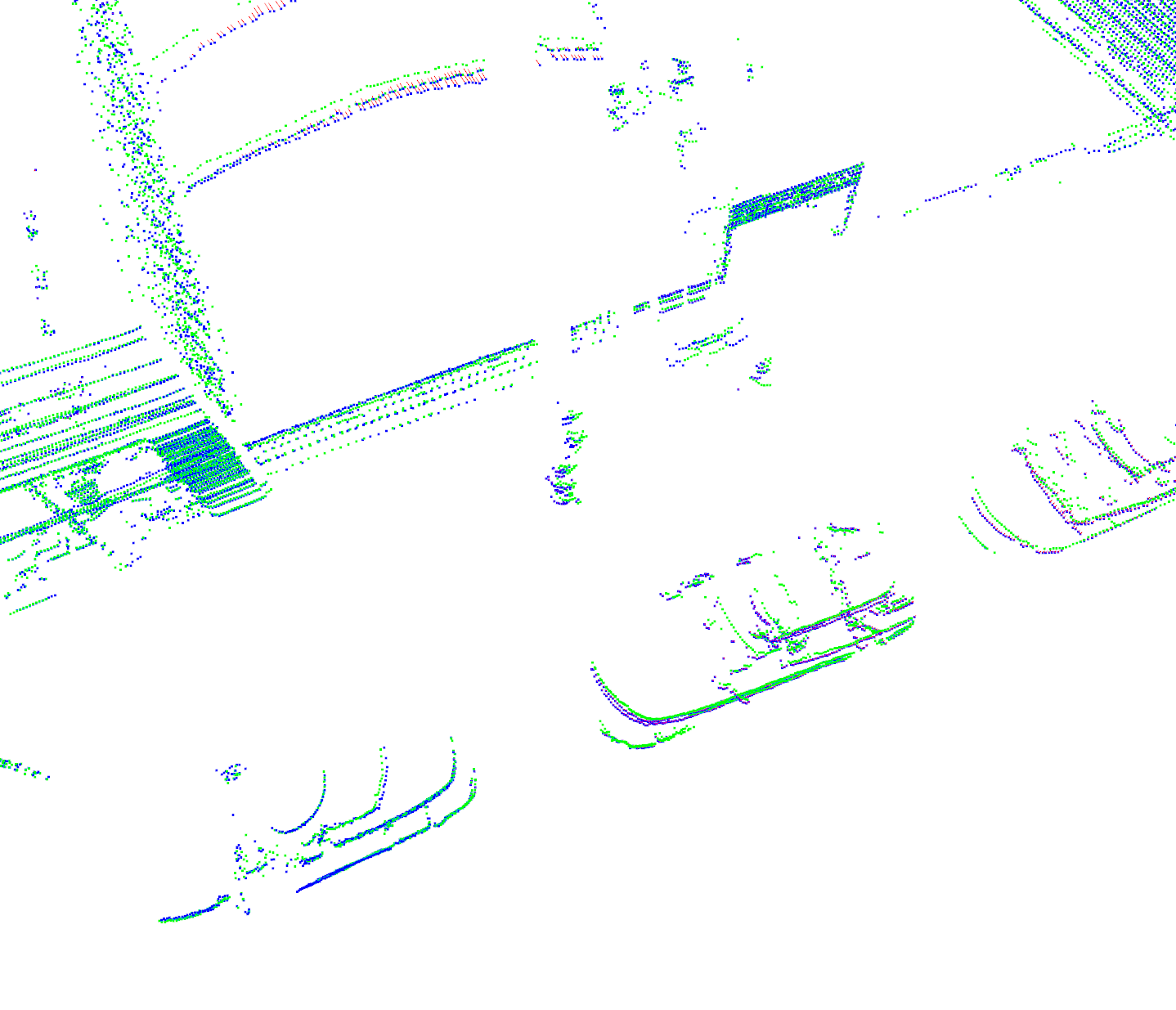}
  \includegraphics[width=\linewidth]{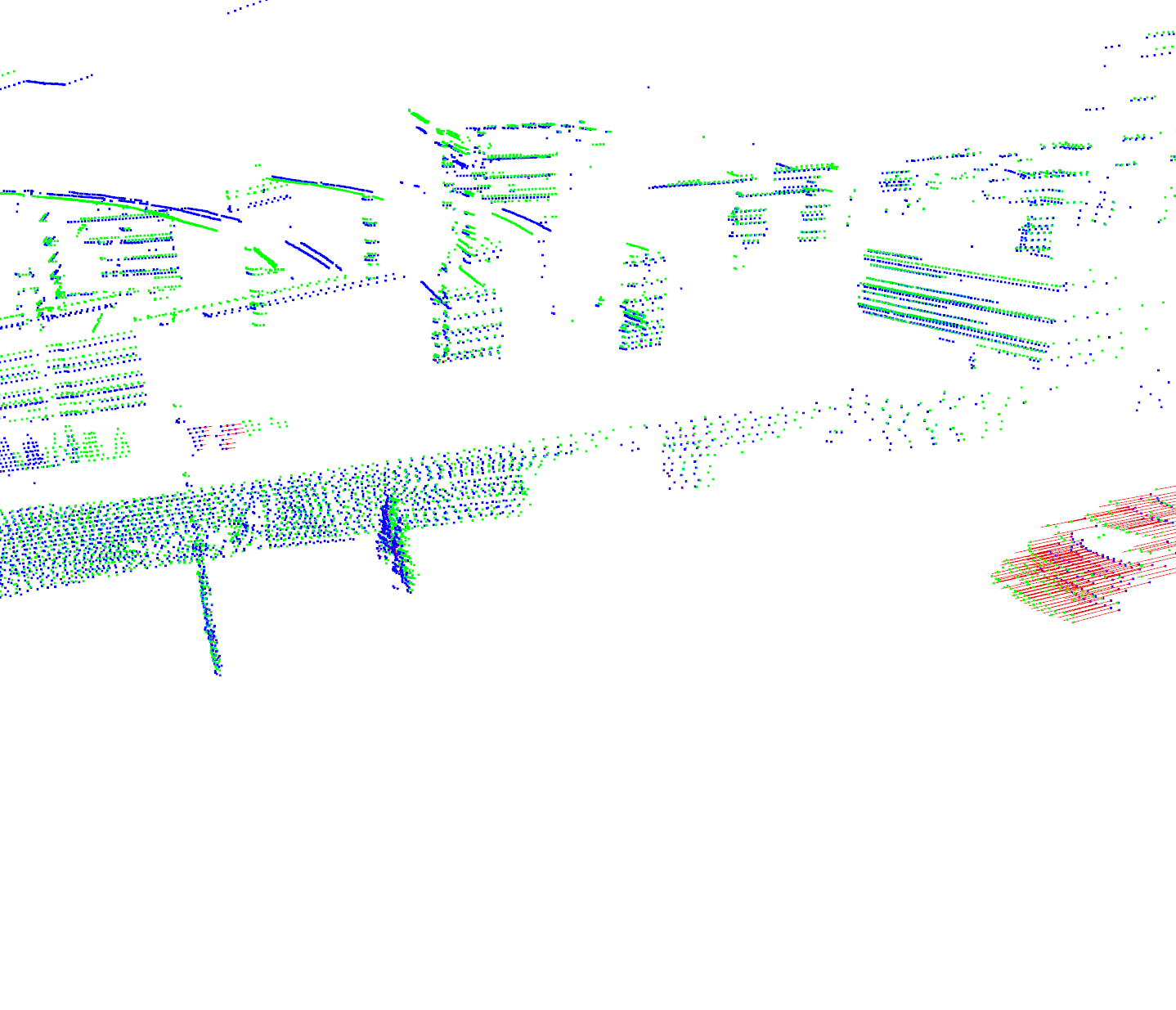}
  \includegraphics[width=\linewidth]{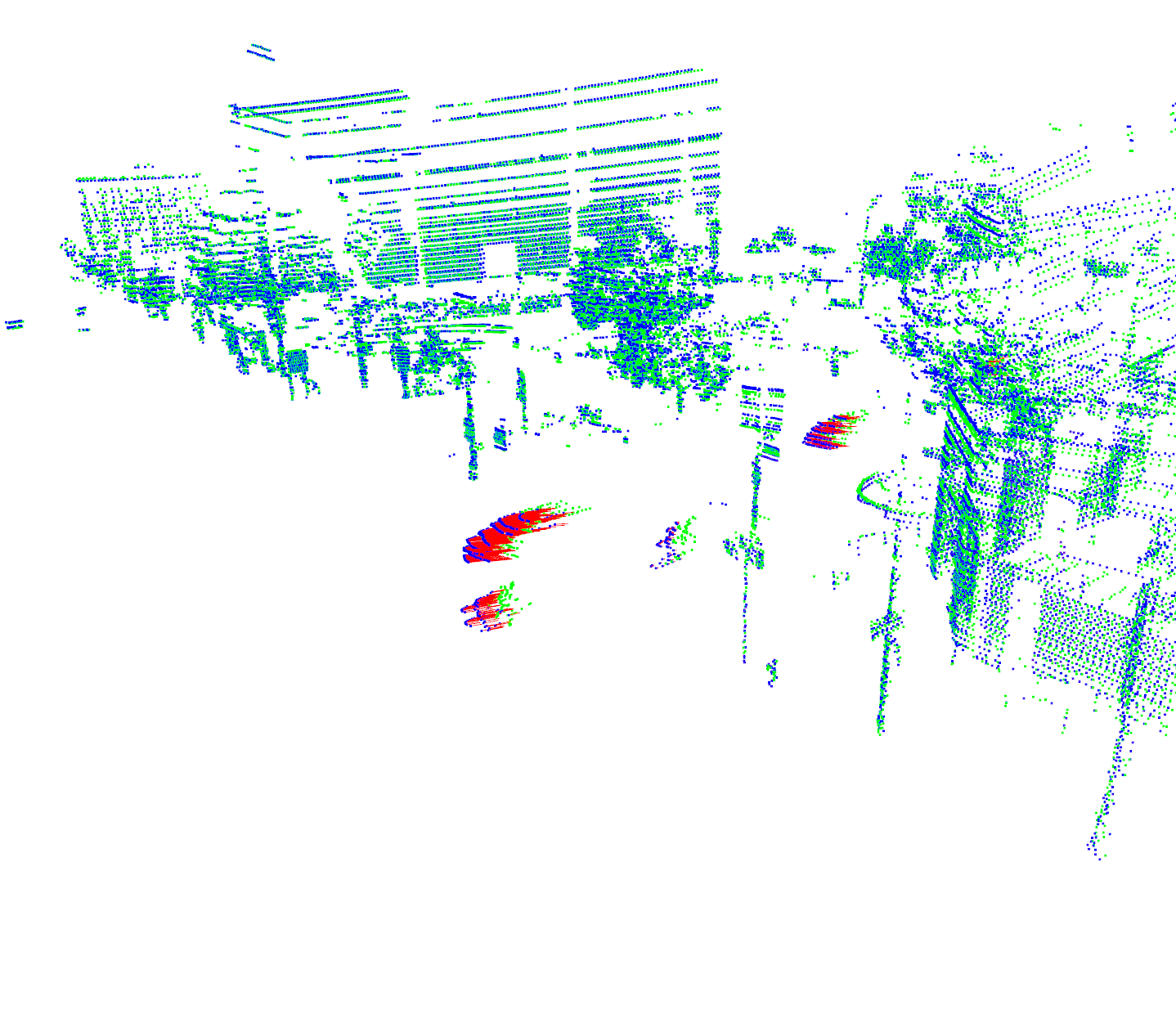}
  \includegraphics[width=\linewidth]{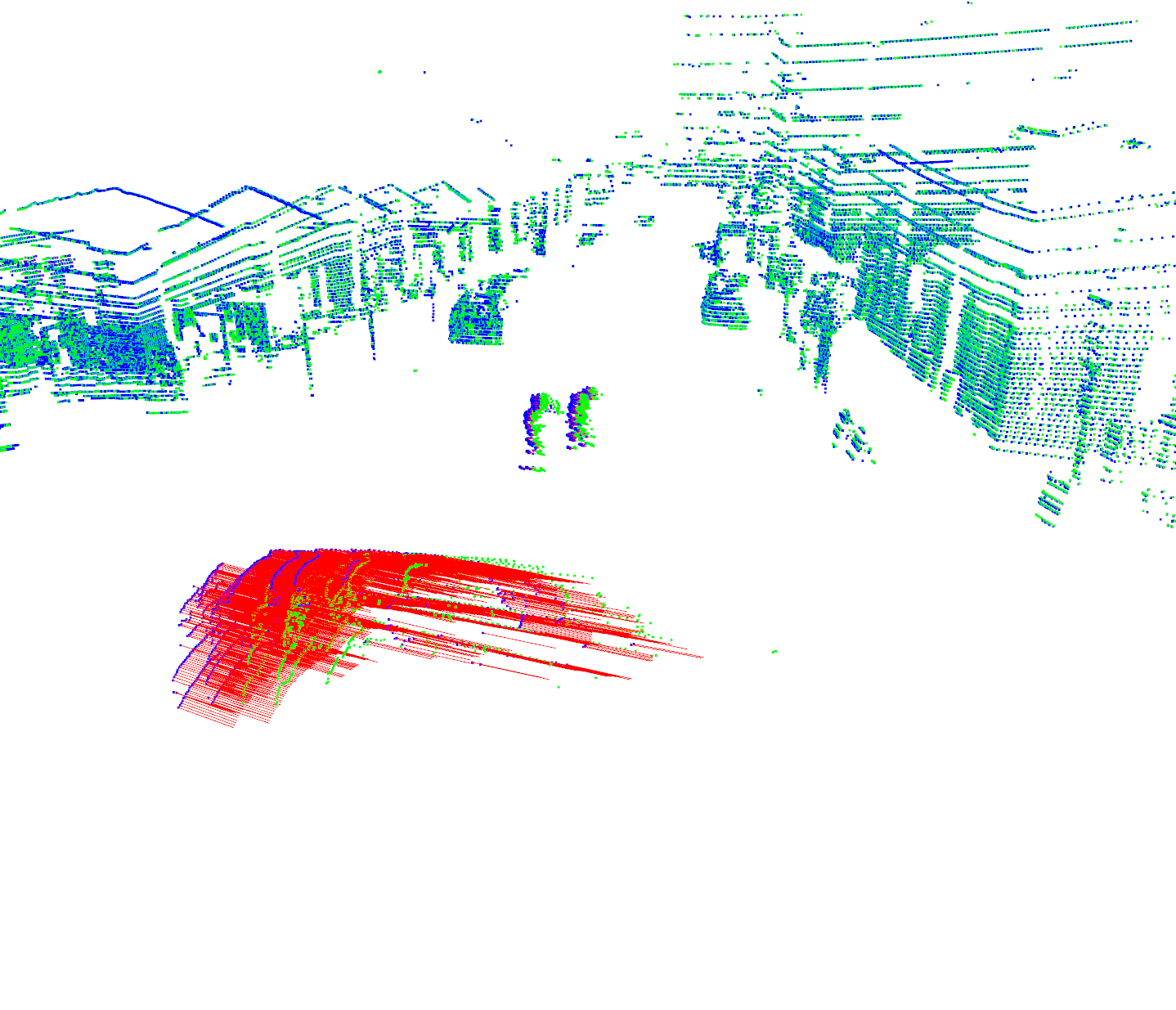}
  \caption{\tinier ZeroFlow}
  \label{fig:sub1}
\end{subfigure}%
\begin{subfigure}{.16\textwidth}
  \centering
  \includegraphics[width=\linewidth]{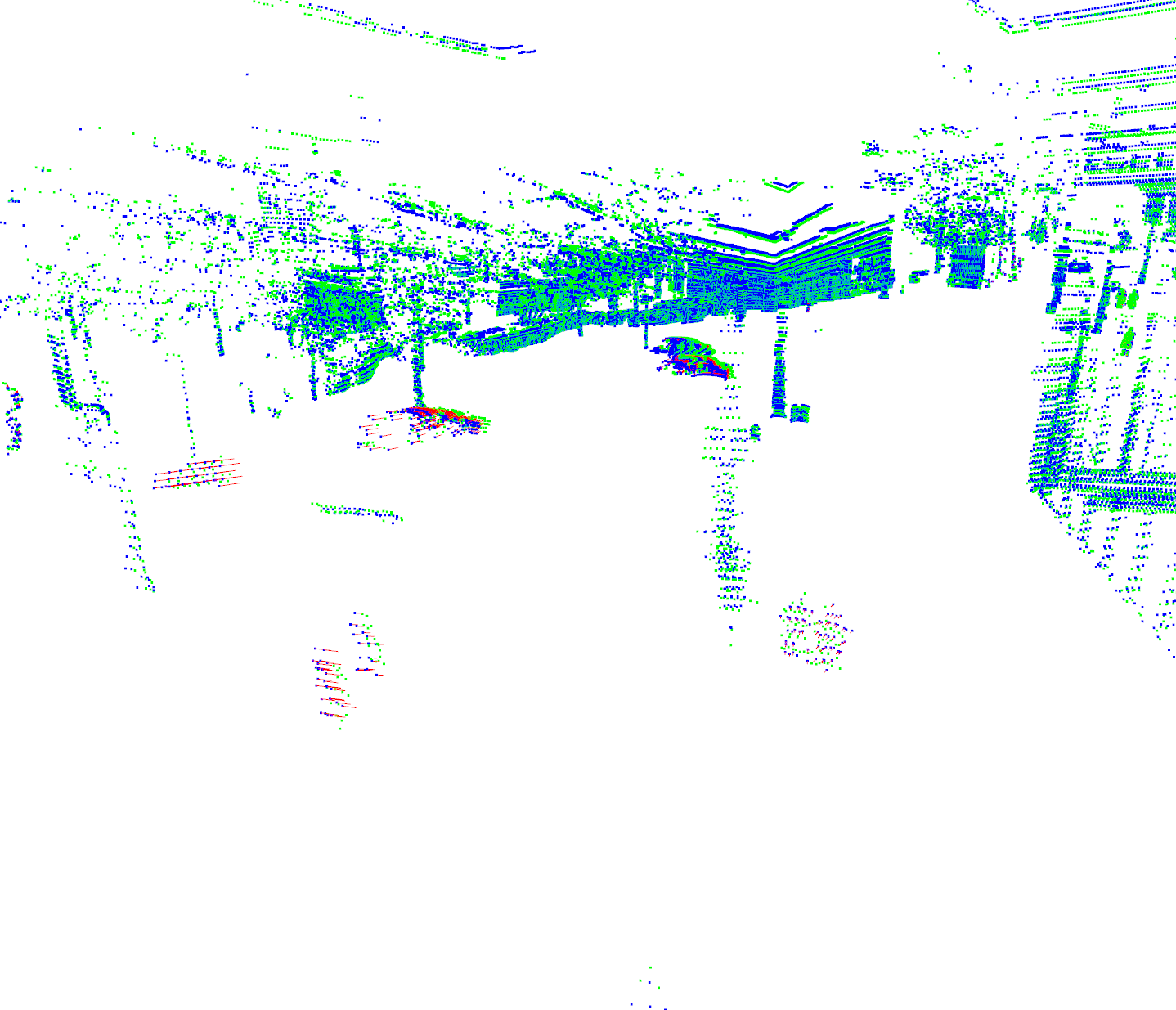}
  \includegraphics[width=\linewidth]{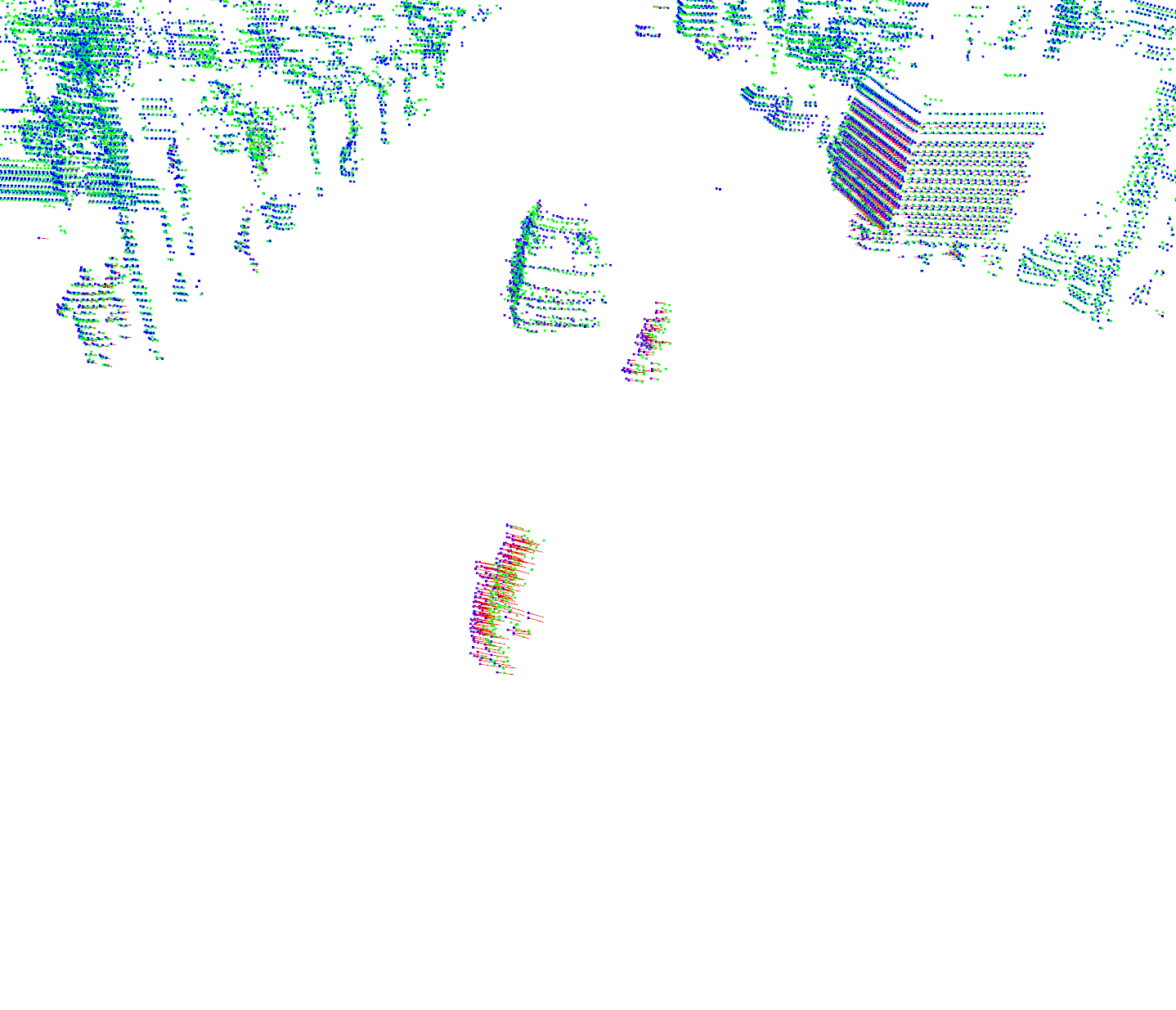}
  \includegraphics[width=\linewidth]{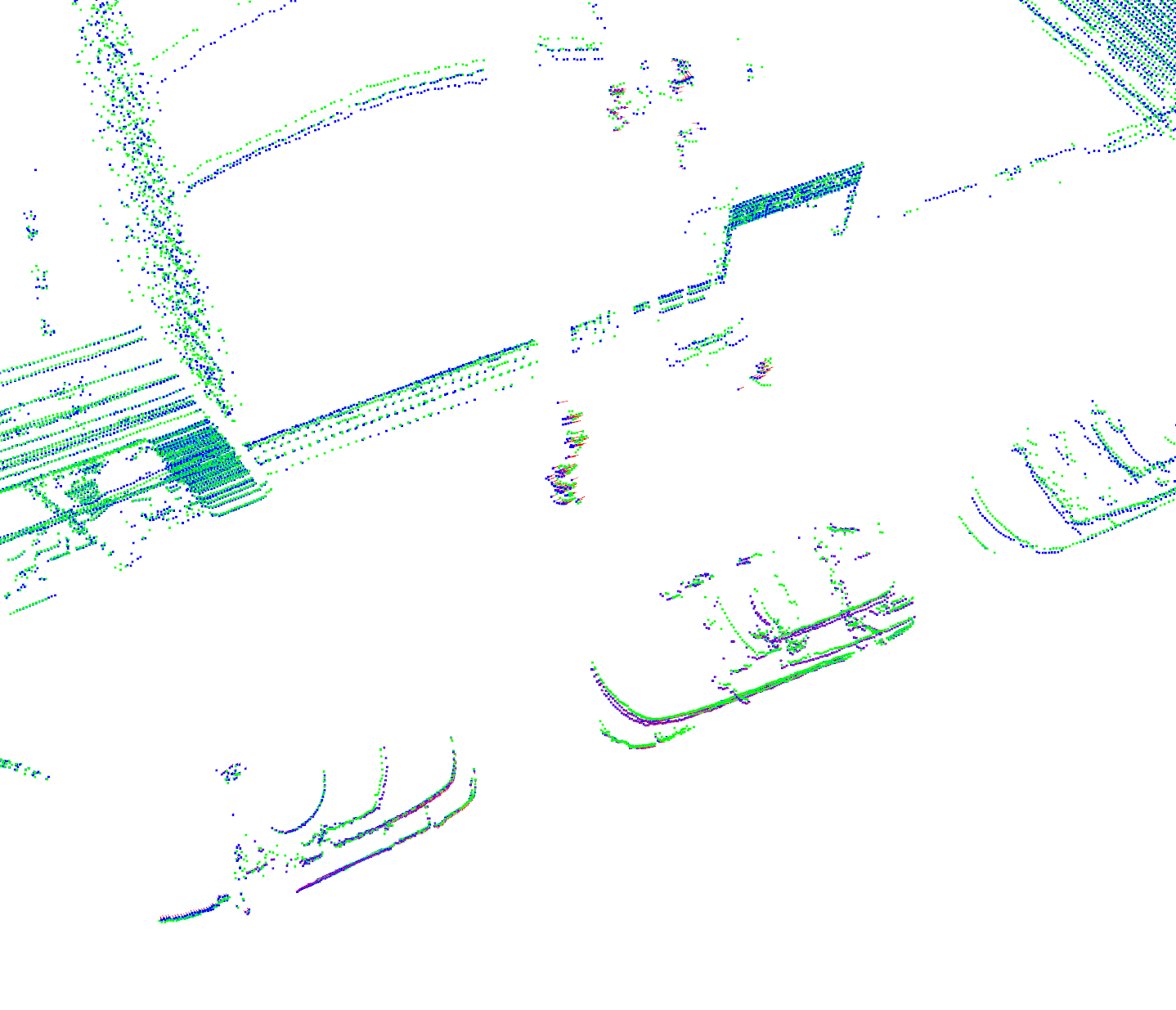}
  \includegraphics[width=\linewidth]{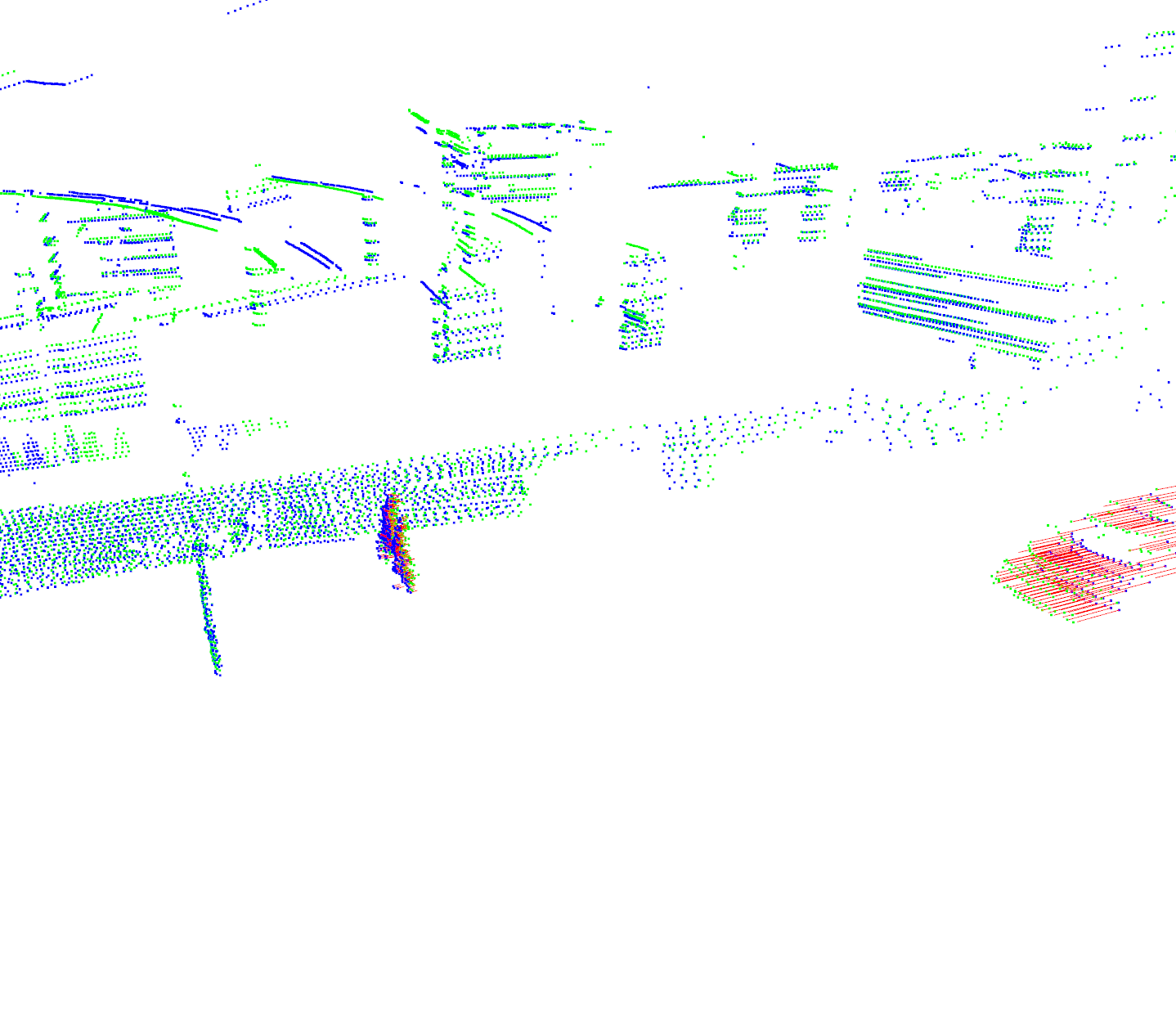}
  \includegraphics[width=\linewidth]{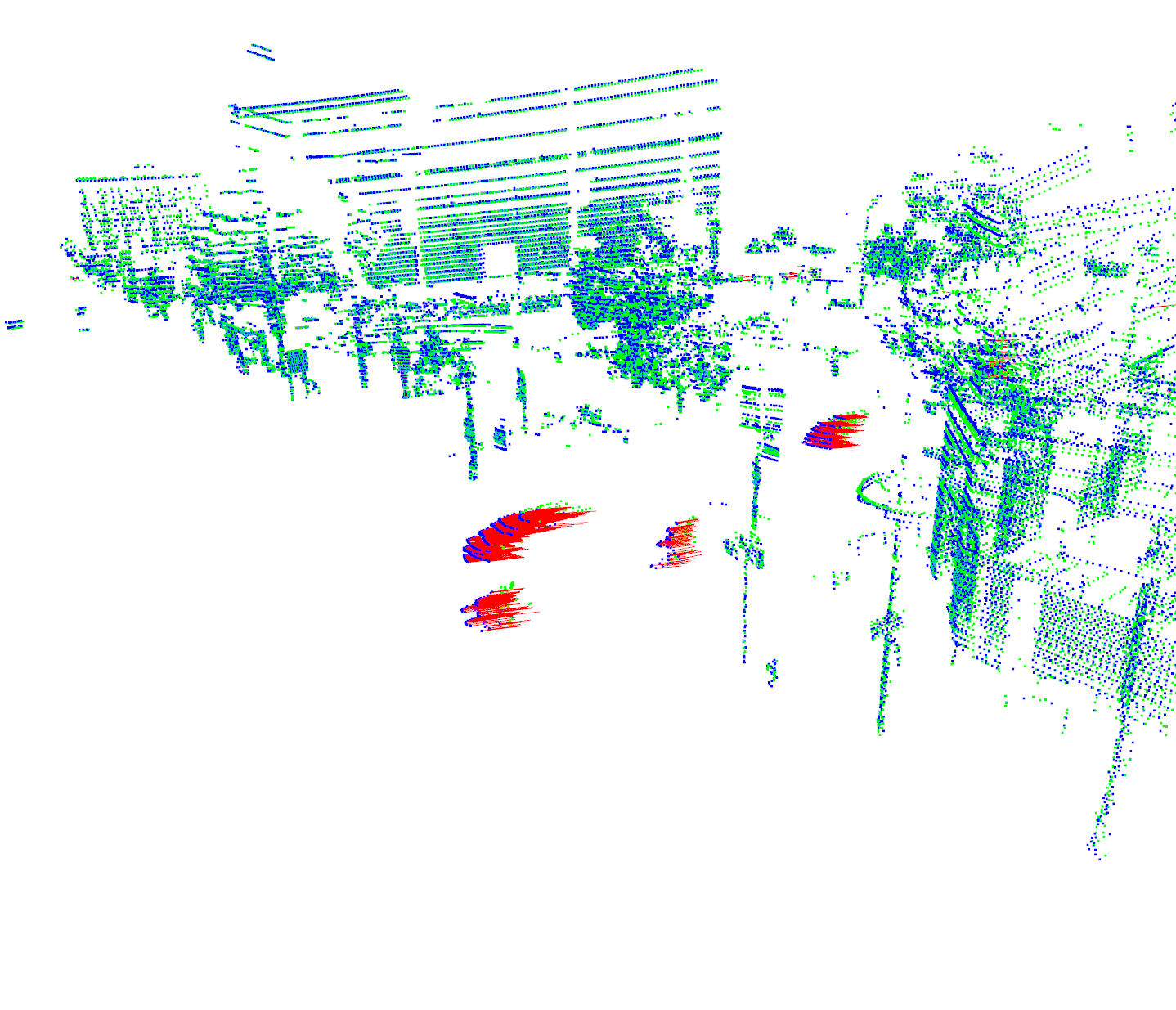}
  \includegraphics[width=\linewidth]{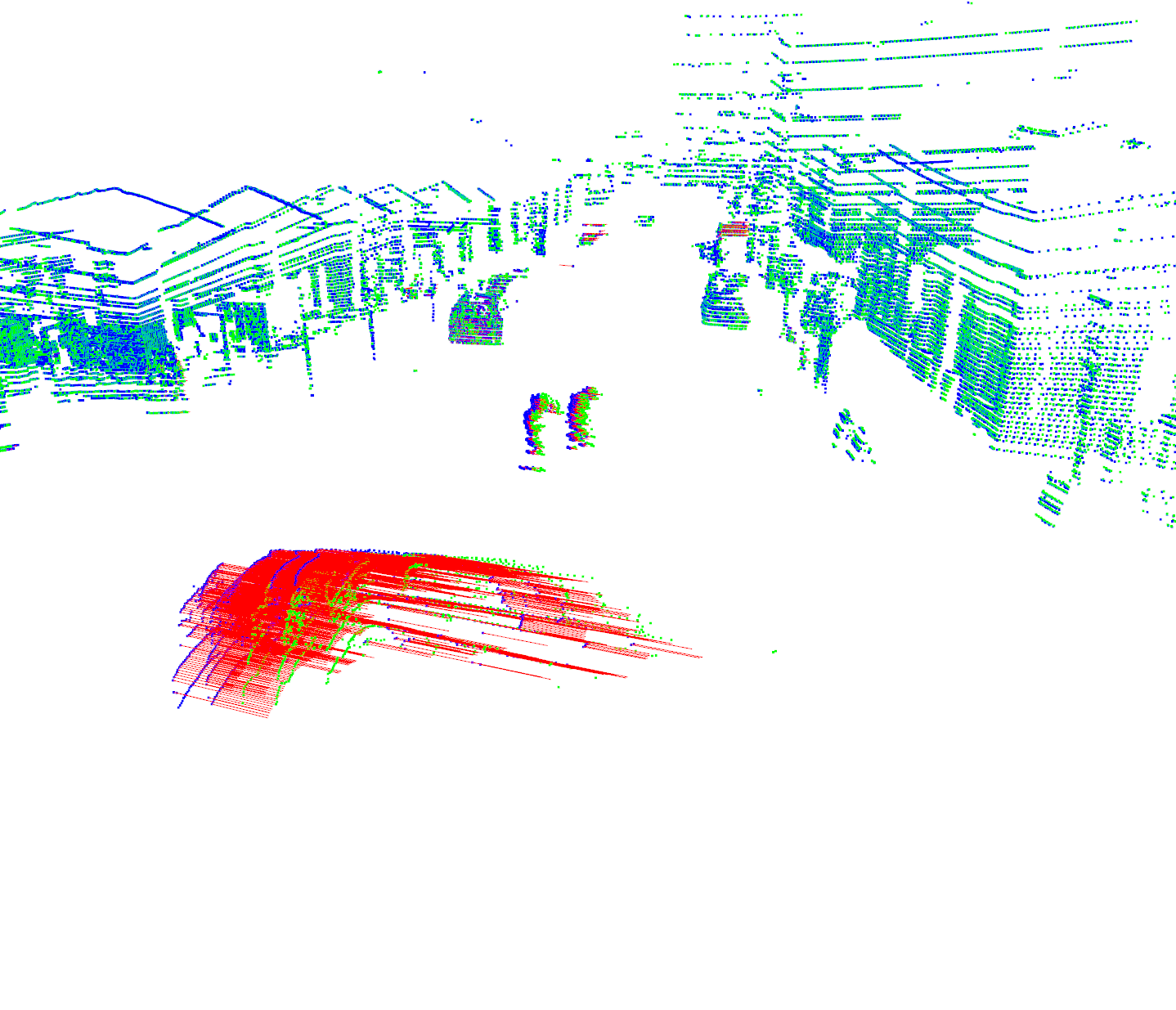}
  \caption{\tinier \textbf{\ourmethod{}}}
  \label{fig:sub6}
\end{subfigure}
\captionsetup{singlelinecheck=off}
\caption[foo bar]{Visualizations of different methods on diverse scenes in Argoverse 2. Each method is estimating flow from the \blue{blue} to the \green{green} point cloud.
\begin{enumerate}[label=Row \arabic*:,leftmargin=4em]
    \item Two pedestrians in left foreground with cars moving in the background. \ourmethod{} is the only method able to describe the pedestrian motion.
    \item Three pedestrians walking across an intersection in front of a stationary car. DeFlow is able to capture the furthest pedestrian, but only \ourmethod{} is able to capture the motion of all three. \ourmethod{} also falsely estimates motion of the moving box truck in the background.
    \item Top view of pedestrians walking down the sidewalk between a building and several cars parked in the street. \ourmethod{} is the only method able to describe the pedestrian motion.
    \item Pedestrians walking down the sidewalk next to a moving car.  \ourmethod{} is the only method able to describe the pedestrian motion.
    \item Two bicyclists riding across an intersection next to driving cars. Most methods are able to capture the training bicyclists and the moving cars, but only NSFP and \ourmethod{} are able to capture the lead bicyclist.
    \item Two pedestrians walk across an intersection while a car drives parallel to them. All methods capture the car motion, but only DeFlow, NSFP, and \ourmethod{} capture most of the pedestrian motion. \ourmethod{} also falsely estimates motion of one of the parked cars far down the street in the background.
\end{enumerate}
}
\figlabel{morequalitativeone}
\end{figure}

\section{Conclusion}

In this work, we highlight that current scene flow methods consistently fail to describe motion on pedestrians and other small objects. We demonstrate that current standard evaluation metrics hide this failure and present \oureval{}, a new class-aware, speed normalized evaluation protocol, to quantify this failure. In addition, we present \ourmethod{}, a frustratingly simple supervised scene flow baseline that achieves state-of-the-art on Threeway EPE and \oureval{}. We argue that current evaluation protocols fail to reveal performance across the distribution of safety-critical objects, and do not contextualize absolute errors in the context of an object's speed. Moreover, we highlight that class and speed aware evaluation is important \emph{even if a method has zero human supervision}. Importantly, we cannot expect any method to meaningfully generalize to the long tail of unknown objects if it cannot provide high quality motion descriptions on a known set of objects. Lastly, \ourmethod{} outperforms prior art by a wide margin because it leverages recent advances in class imbalanced learning. Our approach highlights that supervised scene flow methods should adopt many of the lessons learned by the detection community to properly address class and point imbalances. 

\subsection{Limitations}

\ourmethod{} only predicts rigid flow for objects within LE3DE2E's fixed taxonomy because it uses a closed-world bounding box based detector. However, as discussed in \appendixref{faq}, these limitations can be addressed with a different detector architectures, and is not a fundamental limitations of the \ourframework{} framework.

\noindent\textbf{Acknowledgements:}
This work was supported in part by
funding from the NSF GRFP (Grant No. DGE2140739). This work was in part supported by the Army Research Office under MURI award W911NF20-1-0080. Any opinions, findings, conclusions, or recommendations expressed in this material are those of the authors and do not necessarily reflect the view of the Army or the US government.

\bibliographystyle{splncs04}
\bibliography{references}
\newpage

\appendix

\section{Argoverse 2 2024 Scene Flow Challenge}

\begin{figure}[htb]
\centering
\includegraphics{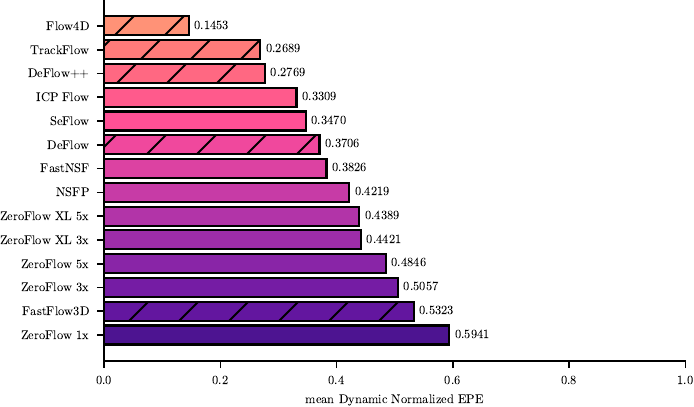}
\caption{mean Dynamic Normalized EPE of submissions to \emph{the Argoverse 2 2024 Scene Flow Challenge} on Argoverse 2's \emph{test} split. Supervised methods shown with hatching. Lower is better.}
\figlabel{competitionmeandynamic}
\end{figure}
\begin{figure}[h!]
\centering
\begin{subfigure}[b]{0.49\textwidth}
    \centering
    \includegraphics{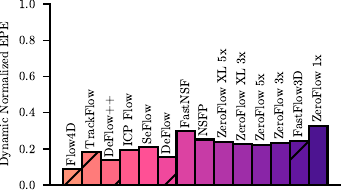}
    \caption{\texttt{CAR}}
    \figlabel{fig:competitioncar}
\end{subfigure}%
\begin{subfigure}[b]{0.49\textwidth}
    \centering
    \includegraphics{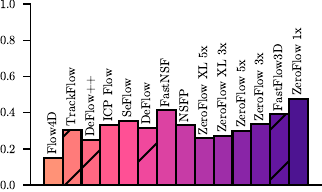}
    \caption{\texttt{OTHER VEHICLES}}
    \figlabel{fig:competitionother-vehicles}
\end{subfigure}
\begin{subfigure}[b]{0.49\textwidth}
    \centering
    \includegraphics{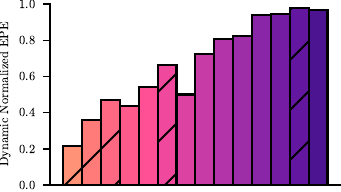}
    \caption{\texttt{PEDESTRIAN}}
    \figlabel{fig:competitionpedestrian}
\end{subfigure}%
\begin{subfigure}[b]{0.49\textwidth}
    \centering
    \includegraphics{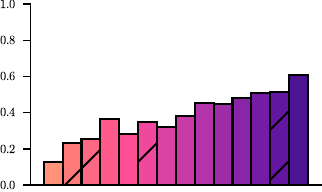}
    \caption{\texttt{WHEELED VRU}}
    \figlabel{fig:competitionwheeled-vru}
\end{subfigure}
\caption{Per meta-class Dynamic Normalized EPE of submissions to \emph{the Argoverse 2 2024 Scene Flow Challenge} on Argoverse 2's \emph{test} split. Supervised methods shown with hatching. Lower is better. Method color and position is consistent between plots.}
\figlabel{competitionmetacatagorydynamic}
\end{figure}

\oureval{} was the basis for the \emph{Argoverse 2 2024 Scene Flow Challenge}\footnote{Full details about the competition can be found at \url{http://argoverse.org/sceneflow}}. The competition featured two tracks: a supervised track, and an unsupervised track, with \ourmethod{} serving as a baseline in the supervised track. Leaderboards for both tracks are ranked by minimum \emph{mean Dynamic} component of \oureval{}.

Notably, Flow4D~\cite{flow4d} significantly improved over all prior supervised methods, halving the dynamic error of \ourmethod{}. Interestingly, Flow4D does not feature any class-aware loss features, instead focusing on architectural improvements over FastFlow3D~\cite{scalablesceneflow}-based architectures (e.g. ZeroFlow~\cite{vedder2024zeroflow}, DeFlow~\cite{zhang2024deflow}). Unsupervised scene flow methods also saw meaningful improvements; ICP-Flow~\cite{lin2024icp} significantly outperformed FastNSF~\cite{fastnsf}, the best performing unsupervised baseline, closely followed by SeFlow~\cite{seflow}.

\section{\oureval{} Structure}\appendixlabel{ourevalstructure}

\tableref{speedclassmatrix} show the structure of the class-speed matrix of \oureval{} on Argoverse 2.

\begin{table}[h!]
\centering
\begin{tabular}{l|ccccc}
\toprule
Class & \multicolumn{5}{c}{Speed Columns} \\
\midrule 
 & 0-0.4m/s & 0.4-0.8m/s & 0.8-1.2m/s & ... & 20-$\infty$m/s \\
\midrule
\texttt{BACKGROUND} & - & - & - & - & - \\
\texttt{CAR} & - & - & - & - & - \\
\texttt{OTHER VEHICLES} & - & - & - & - & - \\
\texttt{PEDESTRIAN} & - & - & - & - & - \\
\texttt{WHEELED VRU} & - & - & - & - & - \\
\bottomrule
\end{tabular}
\caption{Example of \oureval{}'s class-speed matrix.}
\tablelabel{speedclassmatrix}
\end{table}

\section{\oureval{} Without Semantics}\appendixlabel{semanticsfree}

In \sectionref{eval} we present \oureval{} with the object distribution broken down by semantic classes. While this makes sense when semantics are available, this is not a fundamental requirement for \oureval{}. To demonstrate this, we break down Argoverse 2's bounding boxes by \emph{size} instead of semantics. We group the ground truth boxes into one of three volume based clusters: \texttt{SMALL}: $<9.5m^3$, \texttt{MEDIUM}:  $\ge 9.5m^3 \land <40m^3$, or \texttt{LARGE}: $\ge40m^3$. As shows in \figref{semanticsfree}, this distribution breakdown still highlights the poor performance of prior art on small objects. 

\begin{figure}[h!]
\centering
\vspace{-1em}
\begin{subfigure}[b]{0.32\textwidth}
    \centering
    \includegraphics[width=\textwidth]{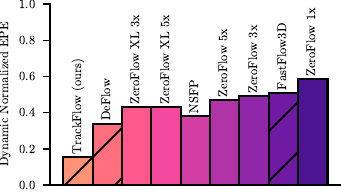}
    \caption{\texttt{SMALL}}
\end{subfigure}%
\begin{subfigure}[b]{0.32\textwidth}
    \centering
    \includegraphics[width=\textwidth]{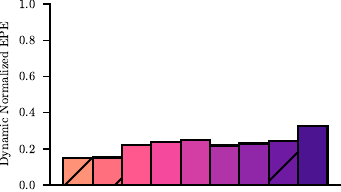}
    \caption{\texttt{MEDIUM}}
\end{subfigure}
\begin{subfigure}[b]{0.32\textwidth}
    \centering
    \includegraphics[width=\textwidth]{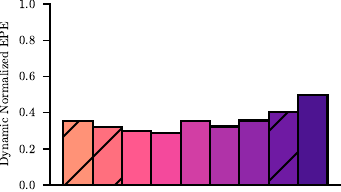}
    \caption{\texttt{LARGE}}
\end{subfigure}%
\caption{\oureval{} using ground truth size based clustering.}
\vspace{-1.5em}
\figlabel{semanticsfree}
\end{figure}

\section{FAQ}\appendixlabel{faq}

\subsection{\ourmethod{} is \emph{just} a tracking method}

Yes, \ourmethod{} is a tracking method applied to the scene flow problem. The state-of-the-art performance of \ourmethod{} suggests that \ourframework{} is a fruitful area of exploration for future work on supervised scene flow.

\subsection{\ourmethod{} uses bounding boxes and thus can only estimate rigid flow --- what does this paper have to say about non-rigid scene flow?}

It's true that \ourmethod{} operates on the level of bounding boxes, but as we discuss in \sectionref{sceneflowdatasets}, public real-world datasets derive motion annotations from bounding box tracks. If non-rigid labels were available, one could train a detector to also regress keypoints (or use an off-the-shelf pretrained method~\cite{yang2023unipose}) and track across those keypoints.

\subsection{\ourmethod{} uses bounding boxes from a detector --- does this mean it cannot detect open-set objects?}

\ourmethod{} uses a class-aware object detector as its bounding box proposer. However, the \ourframework{} framework does not require class annotations -- nothing prevents the use of a class agnostic open world bounding box proposer, either trained like FasterRCNN's RPN~\cite{fastrcnn,fasterrcnn}, Object Localization Network~\cite{kim2021oln}, or via geometric priors~\cite{pmlr-v205-huang23b}.

\subsection{Our metric is ``just'' Threeway EPE extended to multiple classes and multiple speed buckets with normalization, and our method ``just'' combines a detector and tracker. Where is the novelty in this idea?}

The ideas presented in this paper are simple and post-hoc obvious, but serve to highlight catastrophic failures currently overlooked in existing approaches. 

\end{document}